\documentclass[letterpaper]{article}
\usepackage[preprint]{aaai2027}
\usepackage[hyphens]{url}
\usepackage{graphicx}
\usepackage{natbib}
\usepackage{caption}
\usepackage{booktabs}
\usepackage{colortbl}
\definecolor{grouprow}{gray}{0.95}
\usepackage{amsmath}
\usepackage{amssymb}
\usepackage{algorithm}
\usepackage{algpseudocode}
\usepackage[most]{tcolorbox}

\algrenewcommand\algorithmicrequire{\textbf{Input:}}
\algrenewcommand\algorithmicensure{\textbf{Output:}}

\newtcblisting{promptbox}[2][breakable]{
  #1, enhanced, listing only,
  colback=white, colframe=black,
  coltitle=black, colbacktitle=white,
  fonttitle=\bfseries\small, title={#2},
  boxrule=0.5pt, titlerule=0.4pt, arc=1pt, boxsep=1pt,
  left=5pt, right=5pt, top=2pt, bottom=2pt,
  listing options={
    basicstyle=\ttfamily\footnotesize,
    breaklines=true, breakindent=0pt,
    columns=fullflexible, keepspaces=true,
  },
}

\title{GraFT: A Training-Free Framework for Spatial Reasoning in Multimodal Large Language Models via 3D Scene Graphs}

\author{
    Junqing Du,
    Fernando Ropero,
    Erkin Turkoz,
    Yanfeng Zhang\thanks{Corresponding author: zhangyanfeng8@huawei.com},
    Lu Liu
}
\affiliations{
    Riemann Lab, Huawei Technologies
}

\begin{document}

\maketitle

\begin{abstract}
3D spatial reasoning underpins understanding and acting in the physical world, yet it remains unreliable in current multimodal large language models (MLLMs). These models falter at precise geometric measurement, at transforming between egocentric and allocentric viewpoints, and at grounding fine-grained appearance. The most common remedies fine-tune the model on large-scale curated spatial-reasoning datasets or attach dedicated encoders for 3D geometry, which typically couples the solution to costly supervision and a specific backbone. We instead introduce GraFT, a training-free framework that supplies the missing 3D structure through a compact, easily maintained 3D scene graph (3DSG). GraFT supplies the graph's 3D structure to the frozen model in three forms: as values computed by \emph{symbolic} tools, as a \emph{bird's-eye-view} (BEV) the model reads as an image, and as \emph{egocentric} frames the geometry selects. On ScanQA, GraFT improves every metric over the same-backbone baseline, raising CIDEr by 27\%. On VSI-Bench, GraFT improves frozen MLLMs by up to 65\%, surpassing every proprietary and general-purpose open-source baseline, and several prominent fine-tuned spatial models.
\end{abstract}

% Uncomment to advertise code / data on the first page (allowed now that the
% paper is no longer anonymous).
% \begin{links}
%     \link{Code}{https://github.com/your-org/graft}
% \end{links}

\section{Introduction}

Spatial reasoning, the ability to interpret the spatial structure of a 3D scene,
such as the sizes, distances, and spatial relations of the objects it contains,
is a cornerstone of physical intelligence. It underpins
downstream applications from augmented reality and autonomous driving to
embodied AI~\cite{vsibench,spatialvlm}.
Multimodal large language models (MLLMs) have attracted growing interest as a natural-language
interface for these tasks~\cite{vsibench,struct2d}.
However, they still fall markedly
short on spatial reasoning, a gap consistently documented across recent benchmarks
and surveys~\cite{vsibench,spatialvlm,spatialsurvey}.

Prevailing approaches build spatial competence into the model itself.
One line of work trains the MLLM for spatial reasoning via supervised fine-tuning~\cite{spatialvlm,struct2d} or reinforcement learning~\cite{spacer,spatialladder,vilasr,r1zerovsi,ocr}; another feeds
explicit 3D representations such as point clouds through specialized
encoders~\cite{pointllm}.
Both approaches not only require large-scale data curation and training, but also
commit, before the task is seen, to a fixed choice: a spatial prior bound to the
training distribution and backbone, or an encoder tied to one input format.
Yet the evidence a spatial task demands varies widely.
Layout-centric tasks benefit from a global, map-like view of the
scene~\cite{gpt4scene,mapnav}.
Tasks about an object's visual attributes instead align with the pretraining strength of MLLMs
once a suitable 2D view is provided. Exact metric grounding, by contrast, is more reliably obtained by delegating to external geometric tools than by relying on learned priors~\cite{tiger,spacetools}.

\begin{figure*}[t]
\centering
\includegraphics[width=\textwidth]{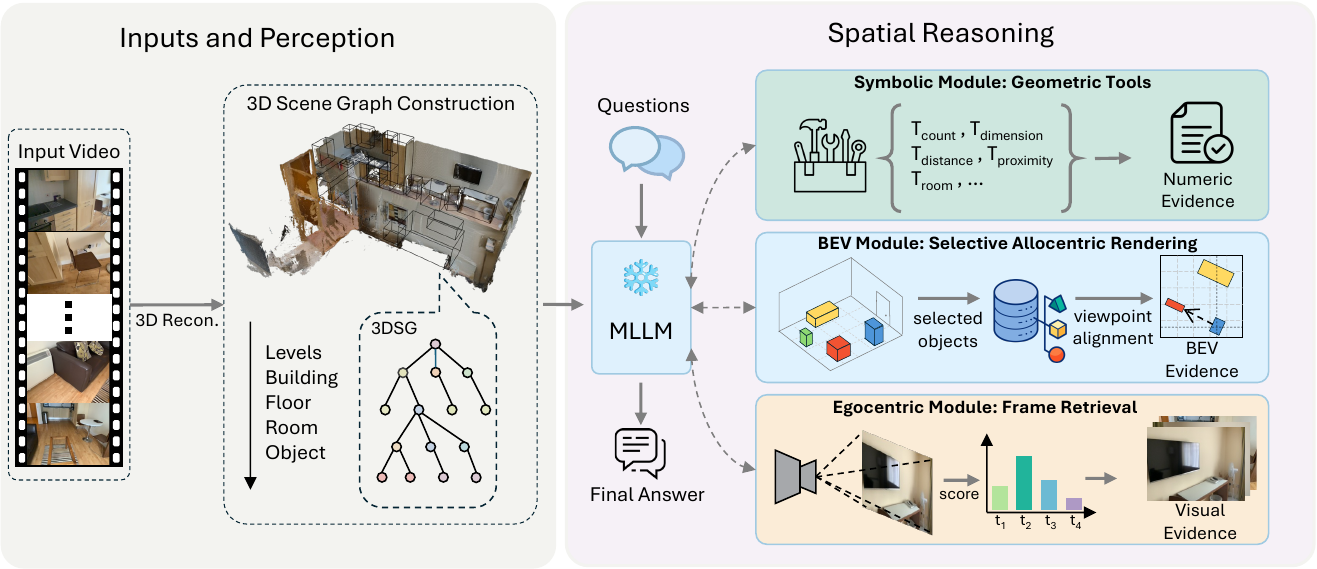}
\caption{Overview of GraFT. Left: an upstream pipeline turns the input video
into a hierarchical 3D scene graph.
Right: the frozen MLLM matches each task to one of three modules
(symbolic tools, BEV rendering, or egocentric frames), which returns evidence
the backbone reasons over to answer, without task-specific training.}
\label{fig:overview}
\end{figure*}

We present GraFT, a training-free framework that instead derives task-appropriate evidence by reasoning through a 3D scene graph (3DSG), a representation the robotics and
embodied-AI community increasingly adopts~\cite{hydra,conceptgraphs,sgnav}.
Unlike raw point clouds, a 3DSG exposes per-object oriented bounding boxes and
semantic labels as a compact symbolic structure that is incrementally
maintainable and readable without a learned encoder.
GraFT treats this graph as the
medium that connects a task to the evidence it demands. Symbolic tools compute exact
answers from the stored 3D attributes, a task-conditioned bird's-eye-view
rendering exposes global layout, and the graph's geometry ranks egocentric
frames by how clearly they show the queried object.
The frozen backbone then reasons over this evidence using two abilities it
already possesses: invoking tools and interpreting structured visual
inputs~\cite{gpt4scene,struct2d}. Each module therefore answers the same
grounding question differently: in what form should 3D structure enter the
model's context, as a number, a rendered view, or a selected frame?

We validate GraFT on two widely used spatial-reasoning benchmarks,
VSI-Bench~\cite{vsibench} and ScanQA~\cite{scanqa}, against strong proprietary,
open-source, and fine-tuned models. On VSI-Bench, GraFT lifts three frozen
open-source backbones by 37\% to 65\% over their standalone performance, and its
best configuration reaches 51.4 on the average. That result is ahead of every
proprietary model (best 45.4) and every general-purpose open-source baseline
(best 40.9) we evaluate, and it surpasses several systems fine-tuned for spatial
reasoning. On ScanQA, replacing uniformly sampled frames with selected
views improves every caption metric under the same backbone, with the largest gain on CIDEr (58.0 to
73.6), all without updating a single parameter.

\noindent Our main contributions are:
\begin{itemize}
  \item \textbf{A training-free framework built on a 3D scene graph.}
    From a single 3DSG, GraFT derives three capabilities
    (symbolic tools, BEV rendering, and egocentric frame selection), giving a frozen MLLM
    3D spatial reasoning with no fine-tuning.
  \item \textbf{A clean, geometric BEV for egocentric-to-allocentric reasoning.} From the scene graph, GraFT renders a top-down view of \emph{only} the task-relevant objects, each drawn as a box with its exact proportions and all other scene content omitted, so the backbone reads the layout from uncluttered, geometrically faithful evidence.
  \item \textbf{A geometry-guided egocentric view retrieval algorithm.}
    Video frames are ranked by the geometric visibility of the queried
    objects, and only the most informative first-person views are forwarded to
    the backbone.
  \item \textbf{Strong spatial reasoning without any training.}
    On VSI-Bench, GraFT surpasses every proprietary and general-purpose open-source baseline, and several models fine-tuned for spatial reasoning; on ScanQA, it improves every metric under the same backbone.
\end{itemize}

\section{Related Work}

\textbf{Spatial reasoning with MLLMs.}
Benchmarks such as ScanQA~\cite{scanqa} and
VSI-Bench~\cite{vsibench} document a persistent gap between the general visual
competence of MLLMs and their spatial understanding.
A common approach augments training: SpatialVLM~\cite{spatialvlm}, SpatialRGPT~\cite{spatialrgpt}, and MM-Spatial~\cite{mmspatial} inject large-scale metric supervision;
SpaceR~\cite{spacer}, SpatialLadder~\cite{spatialladder},
ViLaSR~\cite{vilasr}, R1-Zero-VSI~\cite{r1zerovsi}, and OCR~\cite{ocr} apply
reinforcement learning; and
Spatial-MLLM~\cite{spatialmllm}, VG LLM~\cite{vgllm}, and VLM-3R~\cite{vlm3r} build 3D geometry into the model.
Others instead consume explicit 3D inputs through dedicated encoders, as in
3D-LLM~\cite{hong3dllm}, PointLLM~\cite{pointllm}, and Chat-Scene~\cite{chatscene}.
Both tie spatial competence to training, so a new scene needs retraining; GraFT instead keeps the backbone frozen and changes only its input.

\textbf{Structured visual prompting.}
A complementary line reformats the scene into inputs that MLLMs already parse
well.
Set-of-Mark prompting overlays visual anchors on images~\cite{som}; GPT4Scene pairs video frames with a global BEV image~\cite{gpt4scene}; and
Struct2D~\cite{struct2d} pairs a marked BEV with object metadata, in
both zero-shot and fine-tuned forms.
Two recent training-free methods reformat differently.
See\&Trek~\cite{seetrek} prompts the model from vision alone, sampling
keyframes by semantic richness and encoding relative position into them, without
any 3D input. GR3D~\cite{gr3d} annotates objects with identifiers and encodes
their geometric attributes as text.
Each of these schemes fixes a single form in advance, whether a rendering, a
frame sampler, or a text encoding of geometry. GraFT derives all three from one
3DSG and selects among them per task.

\textbf{Tool-augmented spatial reasoning.}
A further line grants the model access to external tools.
SpatialScore's SpatialAgent~\cite{spatialscore}, TIGeR~\cite{tiger}, and SpaceTools~\cite{spacetools} let an MLLM invoke
estimators for depth, pose, or bounding boxes, typically re-running perception
for each question.
GraFT runs this perception only once, at graph construction; its reasoning-time tools then perform deterministic geometry over the graph, amortizing perception across all questions.

\textbf{3D scene graphs in embodied AI.}
Introduced as a unified structure over semantics, 3D geometry, and
viewpoints~\cite{armeni3dsg}, the 3D scene graph is now widely adopted across
robotics and embodied AI~\cite{sg3dsurvey}. Such graphs are built in real time and with open-vocabulary or hierarchical semantics~\cite{hydra,conceptgraphs,hovsg}, and applied to navigation, embodied QA, and object search, usually by serializing the graph into a textual prompt~\cite{sgnav,grapheqa,momallm}.
GraFT broadens this interface, using the same graph for symbolic tool calls, selective renderings, and visibility-ranked frames.

\section{The GraFT Framework}

\noindent\textbf{Overview.} GraFT is a training-free reasoning layer around a frozen multimodal large language model backbone. Given a 3D scene graph $\mathcal{G}$ of an environment, GraFT supplies the backbone with the evidence it needs for spatial reasoning, without updating any of its parameters (Figure~\ref{fig:overview}). GraFT supplies this evidence in one of three forms, each suited to a different family of tasks: symbolic geometric tools for exact metric answers (Section~\ref{sec:tools}), a task-conditioned bird's-eye-view rendering for egocentric-to-allocentric reasoning (Section~\ref{sec:bev}), and geometry-guided egocentric view retrieval for visual attributes (Section~\ref{sec:ego}). The backbone itself selects among them, in a single text-only pass taken before any evidence is built (Section~\ref{sec:router}).

\textbf{Scene representation.} GraFT reasons over a hierarchical 3D scene
graph produced by an upstream perception pipeline and consumed as given, which
decouples the reasoning layer from how the graph is produced. The graph is formalized as
$\mathcal{G}=(\mathcal{N},\mathcal{E})$. The node set $\mathcal{N}$ unions four
layers: a building node $\mathcal{N}_B$, floor nodes $\mathcal{N}_F$, room nodes
$\mathcal{N}_R$, and object nodes $\mathcal{N}_O$, and the edge set $\mathcal{E}$
records containment between adjacent layers.
Building, floor, and room nodes store a coarse class and box geometry,
while each object node carries the detail the reasoning relies on. Its core is the tuple $o_i=(\ell_i,\mathbf{c}_i,\mathbf{s}_i,\mathbf{R}_i,\mathbf{n}_i,\mathcal{V}_i)$: the class label $\ell_i$, the box center $\mathbf{c}_i=(x_i,y_i,z_i)$ in world meters, the full extents $\mathbf{s}_i=(w_i,d_i,h_i)$ along the box axes, the orientation $\mathbf{R}_i$ recovered from a stored unit quaternion, a unit surface normal $\mathbf{n}_i$, and the set $\mathcal{V}_i$ of video frames in which the object is visible. These frames come from the calibrated video that captured the scene, a posed sequence of frames indexed by $t$: each frame $t$ provides a camera orientation $\mathbf{R}_t$, center $\mathbf{p}_t$, and intrinsics $K_t$, so $\mathcal{V}_i$ is the subset of these frames into which $o_i$'s box projects. 

\subsection{Router: Selecting a Module per Task}
\label{sec:router}

The three modules serve different families of tasks, so GraFT must decide
which kind of evidence a task needs before any of it is built. That decision
is delegated to the backbone itself. Given the question text, together with the
answer options when the task is multiple-choice, GraFT performs a single plaintext
forward pass under a fixed system prompt, which lists the available modules and 
provides an explanation for each one. The output is constrained to a single module 
name. GraFT runs the corresponding module, which builds its evidence
from the 3DSG; the backbone then answers from that evidence alone. 
The system prompt and the full set of module names are given in
Appendix~\ref{sec:router-prompt}.

\subsection{Symbolic Module: Geometric Tools}
\label{sec:tools}

\begin{figure}[t]
\centering
\includegraphics[width=0.90\columnwidth]{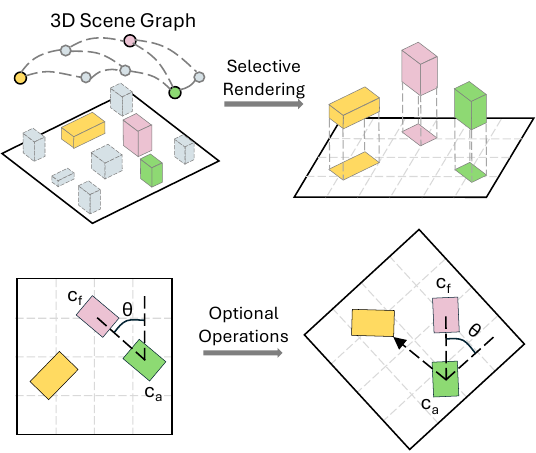}
\caption{Constructing the BEV rendering. Top: from the 3D scene graph, GraFT
renders only the task-relevant objects as box footprints. Bottom: optional
task-conditioned operations rotate the scene about the viewpoint
$\mathbf{c}_a$ by $\theta$ so the facing direction $\mathbf{c}_f$ points up, and
overlay an anchor-to-target arrow.}
\label{fig:bev}
\end{figure}

Each object's box $(\mathbf{c}_i,\mathbf{s}_i,\mathbf{R}_i)$ fixes its position, size, and orientation to the precision of the upstream perception. Any quantity that is a
function of this geometry then follows deterministically from the boxes, so its accuracy is inherited from the perception rather than re-estimated by the backbone. Inferring such quantities from an image would only add error on top. We therefore compute them from the 3DSG directly, through five symbolic
geometric tools that read straight from the boxes, covering \emph{count},
\emph{size}, \emph{distance}, \emph{proximity}, and \emph{room} extent; each is
defined in Appendix~\ref{sec:tools-app}. Each tool runs deterministically over
the boxes, so no part of the numerical computation passes through the backbone.

Resolving a geometric task takes exactly one tool call; GraFT never chains tools or feeds one tool's output into the next.

\subsection{BEV Module: Selective Allocentric Rendering}
\label{sec:bev}

\begin{figure*}[t]
\centering
\includegraphics[width=\textwidth]{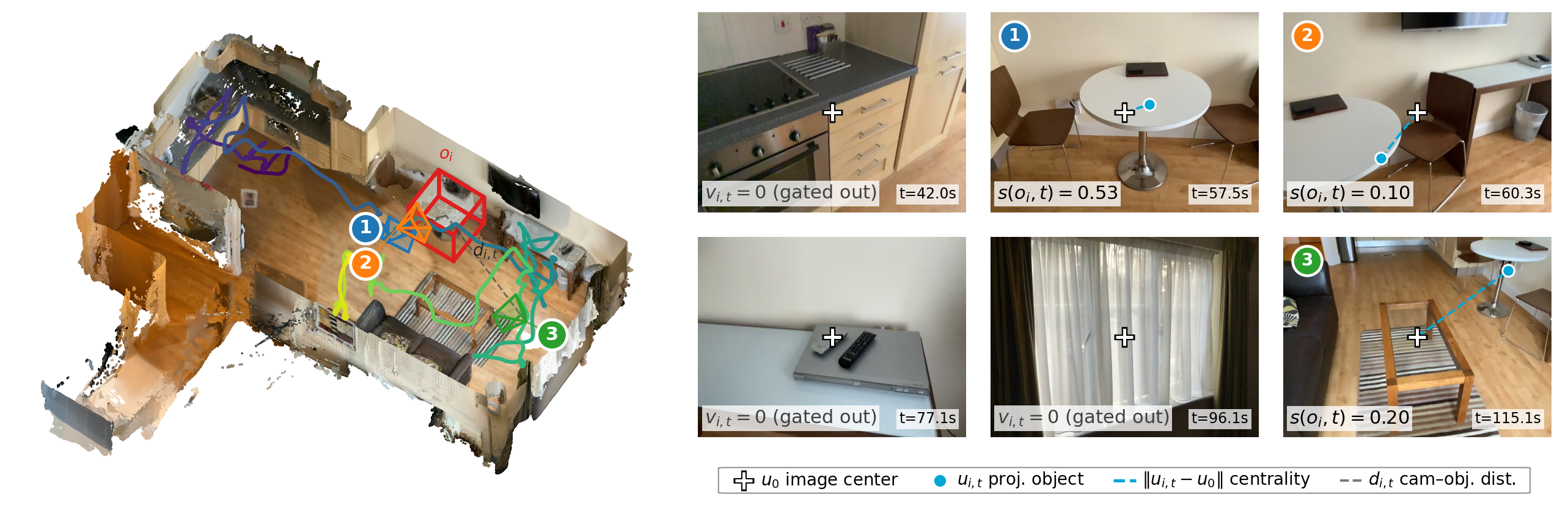}
\caption{Geometry-guided frame scoring for the egocentric module. \emph{Left:} the reconstructed scene with the queried object $o_i$ (red box) and the camera trajectory; the three numbered viewpoints are the frames scored at right, shown at camera--object distance $d_{i,t}$. \emph{Right:} six example frames, shown solely to illustrate the scoring. Each marks the image center $\mathbf{u}_0$ (cross) and the projected object $\mathbf{u}_{i,t}$ (dot), whose offset gives the centrality term $s^{\mathrm{ctr}}$; together with proximity $s^{\mathrm{prox}}$ and corner coverage $s^{\mathrm{cov}}$ they form the visibility-gated score $s(o_i,t)$ (Eq.~(\ref{eq:vis})). Frames in which $o_i$ is not visible are gated out ($v_{i,t}=0$); among the rest, \emph{view~1} ($t{=}57.5$\,s) scores highest ($0.84\times0.63\times1.00\!=\!0.53$: close, centered, fully covered), ahead of \emph{view~3} ($0.56\times0.36\times1.00\!=\!0.20$) and \emph{view~2} ($0.61\times0.65\times0.25\!=\!0.10$, penalized because the table is only partially framed). GraFT forwards the top-scoring frames.}
\label{fig:ego}
\end{figure*}

Reasoning about a scene's global arrangement requires an egocentric-to-allocentric conversion, which a single top-down view makes explicit. GraFT renders such a view directly from the scene graph rather than capturing it with a camera. We build on Struct2D's re-oriented BEV~\cite{struct2d} but render it differently: Struct2D marks the query-relevant objects on a top-down view of the full reconstructed scene, whereas GraFT draws \emph{only} those objects, each as a clean box whose proportions follow its exact dimensions, and omits everything else. The backbone then sees just the layout a task needs, drawn to scale and free of the clutter of a full scene render, which keeps the spatial relations easy to read. Figure~\ref{fig:bev} illustrates this construction.

The viewpoint and heading are likewise set by the task, not by where a camera happened to be. Given a viewpoint center $\mathbf{c}_a$ and a reference point $\mathbf{c}_f$ that fixes the facing direction, GraFT rotates the scene about the viewpoint by an angle $\theta$ that brings the facing direction $\mathbf{c}_f-\mathbf{c}_a$ to the top of the image (Figure~\ref{fig:bev}). The render can additionally carry an optional anchor-to-target arrow that makes a relation explicit rather than leaving it for the backbone to infer. 

\subsection{Egocentric Module: Frame Retrieval}
\label{sec:ego}

Where the symbolic and BEV modules answer from the graph's geometry alone, a task about an object's visual attributes must return to the pixels. For a queried object $o_i$, GraFT scores the posed frames $\mathcal{V}_i$ stored with the object and forwards only the top-$k$ clearest views; Figure~\ref{fig:ego} illustrates the scoring on example frames of one scene.

Whether a frame belongs to $\mathcal{V}_i$ is decided by projecting the object's box into it. Let $v_{i,t}$ denote the visibility of $o_i$ in frame $t$, equal to $1$ when the three geometric conditions of Eq.~(\ref{eq:gate}) hold jointly and $0$ otherwise,
\begin{equation}
\label{eq:gate}
v_{i,t}=\mathbf{1}\!\left[
\begin{array}{c}
n_{i,t}\ge 2,\\[2pt]
d_{\min}\le d_{i,t}\le d_{\max},\\[2pt]
\langle \mathbf{z}_t,\,\mathbf{c}_i-\mathbf{p}_t\rangle\,/\,d_{i,t}\ge\cos\tau
\end{array}
\right].
\end{equation}
In the first condition, $n_{i,t}$ counts how many of the eight box corners project in front of the camera and inside the image, under the frame's perspective projection $\pi(\cdot\,;\theta_t)$. The corners are obtained by offsetting the center $\mathbf{c}_i$ along the box axes $\mathbf{R}_i$ by half the extents $\mathbf{s}_i$ in every sign combination. The calibration $\theta_t=(\mathbf{R}_t,\mathbf{p}_t,K_t)$ collects the camera orientation, center, and intrinsics. Requiring only two corners admits partial views. The second keeps the camera--object distance $d_{i,t}=\|\mathbf{c}_i-\mathbf{p}_t\|$ within an informative depth range. The third requires the camera to face the object within an angle $\tau$: the unit optical axis $\mathbf{z}_t$ is the third column of $\mathbf{R}_t$, so $\langle \mathbf{z}_t,\,\mathbf{c}_i-\mathbf{p}_t\rangle / d_{i,t}$ equals the cosine of the angle between the optical axis and the direction to the object. All thresholds are fixed once and shared across scenes and tasks; the frames that pass constitute $\mathcal{V}_i$. In Figure~\ref{fig:ego}, the frames at $t{=}42.0$, $77.1$, and $96.1$\,s fail these tests and are gated out ($v_{i,t}=0$).

A visible frame is then rated by three terms, each in $[0,1]$. Let $\mathbf{u}_{i,t}$ be the projected center, replaced by the centroid of the in-frame corners when the center falls outside the image; let $\mathbf{u}_0$ be the image center and $\rho$ the image half-diagonal. A centrality term $s^{\mathrm{ctr}}_{i,t}$ rewards a well-framed object,
\begin{equation}
s^{\mathrm{ctr}}_{i,t}=\max\!\Big(0,\;1-\frac{\|\mathbf{u}_{i,t}-\mathbf{u}_0\|}{\rho}\Big),
\end{equation}
peaking when the object sits at the image center; a proximity term $s^{\mathrm{prox}}_{i,t}$ rewards a near object,
\begin{equation}
s^{\mathrm{prox}}_{i,t}=\frac{1}{\max(d_{i,t},\,1)},
\end{equation}
floored at one meter so a very close object does not dominate; and a coverage term $s^{\mathrm{cov}}_{i,t}$ measures how completely the object is framed,
\begin{equation}
s^{\mathrm{cov}}_{i,t}=\frac{n_{i,t}}{8},
\end{equation}
the fraction of box corners inside the image: partial views pass the gate, but fuller framing scores higher. A good view must satisfy all three at once, so the score $s(o_i,t)$ is defined as their product, gated by $v_{i,t}$,
\begin{equation}
\label{eq:vis}
s(o_i,t)=v_{i,t}\,s^{\mathrm{prox}}_{i,t}\,
s^{\mathrm{ctr}}_{i,t}\,s^{\mathrm{cov}}_{i,t},
\end{equation}
and a low value on any one term drags the product down. The three views in Figure~\ref{fig:ego} make this concrete: \emph{view~1} ($t{=}57.5$\,s) is close, centered, and fully covered, and scores highest; \emph{view~3} ($t{=}115.1$\,s) is fully covered but far; and \emph{view~2} ($t{=}60.3$\,s), though near and fairly centered, frames only a quarter of the corners, which alone drags its score down.

Finally, GraFT ranks the frames by the score in~(\ref{eq:vis}), discards near-duplicate viewpoints, and forwards the top-$k$ as the first-person evidence the backbone reasons over.

\section{Experiments}

This section evaluates GraFT on VSI-Bench and ScanQA.
Section~\ref{sec:setup} describes the experimental settings. We then isolate each capability under a fixed backbone: the symbolic
tools on the metric subtasks of VSI-Bench (Section~\ref{sec:exp-tools}), the
BEV rendering on its directional subtasks (Section~\ref{sec:exp-bev}), and the
egocentric retrieval on ScanQA (Section~\ref{sec:exp-ego}). Finally, Section~\ref{sec:exp-real} compares the composed system with prior work, on the VSI-Bench subsets that Struct2D~\cite{struct2d} reports and on the complete benchmark.

\subsection{Experimental Setup}
\label{sec:setup}

\noindent\textbf{Benchmarks.} We use VSI-Bench~\cite{vsibench} and ScanQA~\cite{scanqa}. The per-module
studies use VSI-Bench's five metric subtasks for the symbolic tools, the two
directional subtasks of the 422-question subset Struct2D reports
on~\cite{struct2d} for the BEV, and ScanQA for the egocentric module; the
composed system is evaluated on the complete VSI-Bench. Following the official protocols, VSI-Bench numerical tasks are scored by mean relative accuracy and multiple-choice tasks by exact-match accuracy. ScanQA answers are open-ended text, scored by four standard caption metrics, BLEU, METEOR, ROUGE, and CIDEr, which all measure word and phrase overlap with the reference answers (higher is better).

\noindent\textbf{Backbones.} GraFT is run on frozen, off-the-shelf backbones with no task-specific training: a frontier closed model (GPT-o3) and an open-weight MLLM (Qwen2.5-VL-7B~\cite{qwen25vl}) in the per-module studies (Tables~\ref{tab:tools}--\ref{tab:ego-scanqa}), joined by Qwen2.5-VL-3B and InternVL2-8B~\cite{internvl2} on the complete VSI-Bench (Table~\ref{tab:vsi-realistic}).

\begin{table}[t]
\centering
\small
\setlength{\tabcolsep}{0.6mm}
\begin{tabular}{lcccccc}
\toprule
Method & Avg. & \shortstack{Obj.\\ Cnt.} & \shortstack{Abs.\\ Dist.} & \shortstack{Obj.\\ Size} & \shortstack{Room\\ Size} & \shortstack{Rel.\\ Dist.} \\
\midrule
\multicolumn{7}{l}{\textbf{GPT-o3}} \\
\rowcolor{grouprow}\multicolumn{7}{l}{\textit{Video input}} \\
\quad Video            & 53.2 & 53.1 & 40.6 & 70.0 & 51.4 & 50.9 \\
\cmidrule(lr){1-7}
\rowcolor{grouprow}\multicolumn{7}{l}{\textit{Ground-truth perception}} \\
\quad Serialized Graph & 89.3 & 91.1 & 73.3 & 98.5 & 91.1 & 92.7 \\
\quad GraFT (Ours) & \textbf{95.2} & 89.9 & 97.5 & 99.5 & 91.1 & 98.2 \\
\cmidrule(lr){1-7}
\rowcolor{grouprow}\multicolumn{7}{l}{\textit{Realistic perception}} \\
\quad Serialized Graph & 56.5 & 49.3 & 45.5 & 69.9 & 72.3 & 45.5 \\
\quad GraFT (Ours) & \textbf{58.0} & 49.5 & 43.2 & 66.5 & 83.4 & 47.3 \\
\midrule
\multicolumn{7}{l}{\textbf{Qwen2.5-VL-7B}} \\
\rowcolor{grouprow}\multicolumn{7}{l}{\textit{Video input}} \\
\quad Video            & 36.8 & 33.2 & 15.5 & 51.6 & 43.4 & 40.0 \\
\cmidrule(lr){1-7}
\rowcolor{grouprow}\multicolumn{7}{l}{\textit{Ground-truth perception}} \\
\quad Serialized Graph & 55.4 & 80.0 & 29.7 & 59.2 & 88.3 & 20.0 \\
\quad GraFT (Ours) & \textbf{92.3} & 85.3 & 96.5 & 97.8 & 91.1 & 90.9 \\
\cmidrule(lr){1-7}
\rowcolor{grouprow}\multicolumn{7}{l}{\textit{Realistic perception}} \\
\quad Serialized Graph & 39.3 & 45.0 & 17.9 & 46.9 & 72.3 & 14.5 \\
\quad GraFT (Ours) & \textbf{56.0} & 44.6 & 41.5 & 67.5 & 83.4 & 43.2 \\
\bottomrule
\end{tabular}
\caption{The five metric subtasks of VSI-Bench under three
input forms (raw \emph{Video}, direct reasoning over the 3DSG serialized as text (\emph{Serialized Graph}), and our symbolic tools (\emph{GraFT})), each evaluated with a ground-truth and a realistic-perception 3DSG. Within each perception block the higher Avg.\ of Serialized Graph vs.\ GraFT is in \textbf{bold}.}
\label{tab:tools}
\end{table}

\noindent\textbf{Perception.} We evaluate under three perception settings. (i) \emph{Video input}: the backbone receives the raw video alone, without a 3DSG. (ii) \emph{Ground-truth perception}: the 3DSG is constructed from the ground-truth object annotations of ScanNet~\cite{scannet}, ScanNet++~\cite{scannetpp}, and ARKitScenes~\cite{arkitscenes}. (iii) \emph{Realistic perception}: the 3DSG is instead built by an offline perception pipeline; following prior work~\cite{bundlefusion,struct2d,chatscene}, we reconstruct the scene and detect objects with Mask3D~\cite{mask3d} and UniDet3D~\cite{unidet3d}. Further construction details are provided in Appendix~\ref{sec:impl}.

\subsection{Symbolic Tools for Metric Estimation}
\label{sec:exp-tools}

Table~\ref{tab:tools} asks in what form geometry should reach the backbone:
serialized as text for the model to compute over (\emph{Serialized Graph}), or
computed outside and passed as values (\emph{GraFT}). Both are
compared against the \emph{Video input} on the five VSI-Bench subtasks whose
answers follow from scene geometry, under both perception settings and on two
backbones.

Both \emph{Serialized Graph} and \emph{GraFT} improve over the \emph{Video input},
showing that the 3DSG helps in either form; we then isolate the contrast between
the two. First, \emph{GraFT} is the more accurate on
every setting: on GPT-o3 it improves over \emph{Serialized Graph} by $+5.9$ points
under \emph{ground-truth perception} and $+1.5$ under \emph{realistic perception},
and on Qwen2.5-VL-7B the gains are larger, $+36.9$ and $+16.7$; deterministic
functions (\emph{GraFT}) thus outperform in-context arithmetic (\emph{Serialized
Graph}). Second, \emph{GraFT} makes accuracy far less dependent on the backbone:
under \emph{Serialized Graph} the weaker Qwen2.5-VL-7B trails GPT-o3 by $33.9$
points on \emph{ground-truth perception} and $17.2$ on \emph{realistic perception},
but under \emph{GraFT} it comes within $2.9$ and $2.0$. By computing the answer
over the 3DSG, \emph{GraFT} lets a weaker backbone nearly match a stronger one.

\subsection{BEV Rendering for Egocentric-to-Allocentric Reasoning}
\label{sec:exp-bev}

Table~\ref{tab:bev-vsibench} evaluates the BEV module, whose single
task-conditioned rendering turns an egocentric video into one allocentric view
for reasoning. We test this on two directional VSI-Bench subtasks, relative
direction and route planning, using the same questions and scenes as Struct2D~\cite{struct2d} for
a fair comparison; all methods use GPT-o3, and the table groups them by the
perception settings of Section~\ref{sec:setup}.

GraFT surpasses the \emph{Video input} baseline by $+35.2$ under \emph{ground-truth perception} and $+15.8$ under \emph{realistic perception}. GraFT also leads the prior BEV methods on both settings: its average is $+3.6$ over Struct2D under \emph{ground-truth perception}, and $+18.1$ over GPT4Scene~\cite{gpt4scene} and $+3.3$ over Struct2D under \emph{realistic perception}. This consistent lead over prior BEV methods comes from GraFT's selectivity: it renders only the query-relevant objects and omits the rest, giving the frozen backbone an uncluttered view of the layout a directional question needs.

\noindent\textbf{What makes the rendering work.}
Table~\ref{tab:bev-ablation} isolates the three elements that separate GraFT's
BEV from a plain top-down render, holding the backbone (GPT-o3), the questions,
and the ground-truth 3DSG fixed so that only the render changes. Each element
contributes, but selection is decisive: restricting the render to the
question-relevant objects, rather than drawing every object the 3DSG records
for the scene, accounts for $+38.5$ points, an order of magnitude beyond the
$+4.0$ contributed by rotation and the $+0.8$ by the anchor-to-target arrow. The gain comes from a property of the
3DSG itself, because the graph indexes the scene by object, the
render can omit everything the question does not ask about.

\begin{table}[t]
\centering
\small
\setlength{\tabcolsep}{0.8mm}
\begin{tabular}{lccc}
\toprule
Method & Avg. & Rel.\ Dir. & Route Plan \\
\midrule
\rowcolor{grouprow}\multicolumn{4}{l}{\textit{Video input}} \\
VSI-Bench~\cite{vsibench} & 55.7 & 49.4 & 61.9 \\
\cmidrule(lr){1-4}
\rowcolor{grouprow}\multicolumn{4}{l}{\textit{Ground-truth perception}} \\
Struct2D~\cite{struct2d}    & 87.3 & 94.4 & 80.1 \\
GraFT (Ours)                & \textbf{90.9} & \textbf{96.1} & \textbf{85.7} \\
\cmidrule(lr){1-4}
\rowcolor{grouprow}\multicolumn{4}{l}{\textit{Realistic perception}} \\
GPT4Scene~\cite{gpt4scene}  & 53.4 & 47.9 & 58.8 \\
Struct2D~\cite{struct2d}      & 68.2 & 60.1 & \textbf{76.2} \\
GraFT (Ours)  & \textbf{71.5} & \textbf{71.6} & 71.4 \\
\bottomrule
\end{tabular}
\caption{Zero-shot GPT-o3 on the two directional subtasks of the VSI-Bench subset that Struct2D reports on; GraFT answers with a single allocentric BEV rendering.
Avg.\ is the mean of the two subtasks; baseline numbers are from
Struct2D~\cite{struct2d}. Best per column within each group is in \textbf{bold}.}
\label{tab:bev-vsibench}
\end{table}

\begin{table}[t]
\centering
\small
\setlength{\tabcolsep}{2mm}
\begin{tabular}{cccc}
\toprule
Selection & Rotation & Arrow & Rel.\ Dir. \\
\midrule
--         & --         & --         & 52.8 \\
\checkmark & --         & --         & 91.3 \\
\checkmark & \checkmark & --         & 95.3 \\
\checkmark & \checkmark & \checkmark & \textbf{96.1} \\
\bottomrule
\end{tabular}
\caption{Cumulative ablation of the BEV rendering on the relative-direction
subtask. \emph{Selection} draws only
the question-relevant objects instead of every object the 3DSG records for the
scene; \emph{Rotation} turns the render by $\theta$ so the observer's facing
direction points up; \emph{Arrow} overlays the anchor-to-target arrow.}
\label{tab:bev-ablation}
\end{table}

\begin{table}[t]
\centering
\small
\setlength{\tabcolsep}{1mm}
\begin{tabular}{lccccc}
\toprule
Model & B-1 & B-4 & METEOR & ROUGE & CIDEr \\
\midrule
\rowcolor{grouprow}\multicolumn{6}{l}{\textit{Zero-shot baselines}} \\
InternVL2-8B   & 23.9 & 3.3 & 14.5 & \underline{34.3} & \underline{62.5} \\
MiniCPM-V-2.6  & 25.1 & \underline{8.4} & 11.8 & 31.5 & 60.1 \\
Qwen2.5-VL-7B  & 22.2 & 7.4 & \underline{16.6} & 30.8 & 58.0 \\
\cmidrule(lr){1-6}
\rowcolor{grouprow}\multicolumn{6}{l}{\textit{Fine-tuned}} \\
Chat-3D        & \underline{29.1} & 6.4 & 11.9 & 28.5 & 53.2 \\
\cmidrule(lr){1-6}
\rowcolor{grouprow}\multicolumn{6}{l}{\textit{Training-free (Ours)}} \\
GraFT (Ours) & \textbf{34.2} & \textbf{9.8} & \textbf{21.7} & \textbf{38.0} & \textbf{73.6} \\
\bottomrule
\end{tabular}
\caption{The egocentric module on ScanQA, with a frozen Qwen2.5-VL-7B backbone.
GraFT (Ours) uses the top-2 frames from the realistic-perception 3DSG.
Comparisons: zero-shot InternVL2-8B~\cite{internvl2} and
MiniCPM-V-2.6~\cite{minicpmv}, the fine-tuned Chat-3D~\cite{chat3d}, and the same
Qwen2.5-VL-7B backbone with uniform sampling. B-1/B-4 denote BLEU-1/BLEU-4. Best per column in \textbf{bold};
second best \underline{underlined}.}
\label{tab:ego-scanqa}
\end{table}

\subsection{Egocentric Frame Retrieval on ScanQA}
\label{sec:exp-ego}

Table~\ref{tab:ego-scanqa} evaluates the egocentric module, which grounds attribute questions in pixels: for a queried object, it forwards only the top-$k$ frames that best reveal it to a frozen Qwen2.5-VL-7B backbone. We test on ScanQA~\cite{scanqa}, whose answers often hinge on an object's visual attributes, against zero-shot MLLMs (InternVL2-8B, MiniCPM-V-2.6, and Qwen2.5-VL-7B with uniform sampling) and the fine-tuned Chat-3D.

On the same Qwen2.5-VL-7B backbone, GraFT improves every metric over uniform frame sampling, most notably raising CIDEr from $58.0$ to $73.6$ and BLEU-1 from $22.2$ to $34.2$ (a $+15.6$ and $+12.0$ point gain). Across the full table it also ranks first on all five metrics, ahead of both the zero-shot general-purpose MLLMs and the fine-tuned Chat-3D. Appendix~\ref{sec:kablation} ablates the frame budget $k$, where even a single selected frame beats uniform eight-frame sampling.

\begin{table*}[!t]
\centering
\small
\setlength{\tabcolsep}{1mm}
\begin{tabular}{l|cc|cccccccc}
\toprule
\multicolumn{3}{c}{} & \multicolumn{4}{c}{Numerical Answer} & \multicolumn{4}{c}{Multiple-Choice} \\
\cmidrule(lr){4-7} \cmidrule(lr){8-11}
Method                 & Rank & Avg. & Obj.\ Cnt. & Abs.\ Dist. & Obj.\ Size & Room Size & Rel.\ Dist. & Rel.\ Dir. & Route Plan & App.\ Ord. \\
\hline
\rowcolor{grouprow}\multicolumn{1}{l|}{\textit{Proprietary Models (API)}} & \multicolumn{2}{c|}{} & \multicolumn{8}{l}{} \\
GPT-4o & 18 & 34.0 & 46.2 & 5.3 & 43.8 & 38.2 & 37.0 & 41.3 & 31.5 & 28.5 \\
Gemini-1.5 Flash & 11 & 42.1 & 49.8 & 30.8 & 53.5 & 54.4 & 37.7 & 41.0 & 31.5 & 37.8 \\
Gemini-1.5 Pro & 9 & 45.4 & 56.2 & 30.9 & \underline{64.1} & 43.6 & \textbf{51.3} & \underline{46.3} & \underline{36.0} & 34.6 \\
\hline
\rowcolor{grouprow}\multicolumn{1}{l|}{\textit{Open-source Models}} & \multicolumn{2}{c|}{} & \multicolumn{8}{l}{} \\
InternVL2-8B & 15 & 37.5 & 31.3 & 29.0 & 48.9 & 44.2 & 38.0 & 33.4 & 28.9 & 46.4 \\
InternVL2-40B & 16 & 37.0 & 41.3 & 26.2 & 48.2 & 27.5 & 47.6 & 32.7 & 27.8 & 44.7 \\
VILA-1.5-8B & 23 & 28.9 & 17.4 & 21.8 & 50.3 & 18.8 & 32.1 & 34.8 & 31.0 & 24.8 \\
VILA-1.5-40B & 21 & 31.2 & 22.4 & 24.8 & 48.7 & 22.7 & 40.5 & 25.7 & 31.5 & 32.9 \\
LongVILA-8B & 25 & 21.6 & 29.1 & 9.1 & 16.7 & 0.0 & 29.6 & 30.7 & 32.5 & 25.5 \\
LongVA-7B & 22 & 29.2 & 38.0 & 16.6 & 38.9 & 22.2 & 33.1 & 43.3 & 25.4 & 15.7 \\
Qwen2.5-VL-3B & 24 & 28.7 & 14.6 & 24.3 & 22.1 & 35.0 & 34.1 & 45.7 & 29.4 & 24.3 \\
Qwen2.5-VL-7B & 19 & 33.0 & 40.9 & 14.8 & 43.4 & 10.7 & 38.6 & 38.5 & 33.0 & 29.8 \\
LLaVA-NeXT-Video-7B & 17 & 35.6 & 48.5 & 14.0 & 47.8 & 24.2 & 43.5 & 42.4 & 34.0 & 30.6 \\
LLaVA-NeXT-Video-72B & 12 & 40.9 & 48.9 & 22.8 & 57.4 & 35.3 & 42.4 & 36.7 & 35.0 & 48.6 \\
LLaVA-OneVision-7B & 20 & 32.4 & 47.7 & 20.2 & 47.4 & 12.3 & 42.5 & 35.2 & 29.4 & 24.4 \\
LLaVA-OneVision-72B & 14 & 40.2 & 43.5 & 23.9 & 57.6 & 37.5 & 42.5 & 39.9 & 32.5 & 44.6 \\
\hline
\rowcolor{grouprow}\multicolumn{1}{l|}{\textit{Spatial Reasoning Models}} & \multicolumn{2}{c|}{} & \multicolumn{8}{l}{} \\
VG LLM-8B~\cite{vgllm} & 2 & 50.7 & \textbf{67.9} & 37.7 & 58.6 & 62.0 & 46.6 & 40.7 & 32.4 & \textbf{59.2} \\
Spatial-MLLM~\cite{spatialmllm} & 4 & 48.4 & \underline{65.3} & 34.8 & 63.1 & 45.1 & 41.3 & 46.2 & 33.5 & 46.3 \\
OCR~\cite{ocr} & 5 & 47.5 & 63.2 & 34.1 & 57.4 & 46.7 & 39.6 & 45.5 & \textbf{44.3} & 49.8 \\
SpatialLadder~\cite{spatialladder} & 7 & 45.7 & 63.5 & 34.3 & 61.7 & 43.9 & 45.4 & 44.8 & 35.6 & 36.4 \\
SpaceR~\cite{spacer} & 8 & 45.5 & 57.8 & 28.2 & 59.9 & 47.1 & 40.1 & 45.4 & 33.5 & \underline{52.1} \\
ViLaSR~\cite{vilasr} & 9 & 45.4 & 63.5 & 34.4 & 60.6 & 30.9 & \underline{48.9} & 45.2 & 30.4 & 49.2 \\
R1-Zero-VSI~\cite{r1zerovsi} & 13 & 40.7 & 59.9 & 29.6 & 50.8 & 48.3 & 35.4 & 35.6 & 34.0 & 31.5 \\
\hline
\rowcolor{grouprow}\multicolumn{1}{l|}{\textit{Training-free, Realistic Perception}} & \multicolumn{2}{c|}{} & \multicolumn{8}{l}{} \\
GraFT (Qwen2.5-VL-3B, Ours) & 5 & 47.5 & 46.3 & \underline{42.9} & 62.2 & \underline{83.0} & 41.7 & 31.7 & 25.8 & 46.6 \\
GraFT (Qwen2.5-VL-7B, Ours) & 3 & 50.0 & 44.6 & 42.3 & \textbf{67.5} & \textbf{83.4} & 43.4 & 42.8 & 28.9 & 46.8 \\
GraFT (InternVL2-8B, Ours) & 1 & 51.4 & 46.9 & \textbf{44.1} & 63.8 & \textbf{83.4} & 45.1 & \textbf{46.4} & 31.4 & 49.8 \\
\bottomrule
\end{tabular}
\caption{Comparison on the full VSI-Bench~\cite{vsibench}. GraFT is
training-free and evaluated under realistic perception. Baseline
scores follow the literature (GPT-4o~\cite{gpt4o}; Gemini-1.5~\cite{gemini15};
InternVL2~\cite{internvl2}; VILA-1.5~\cite{vila}; LongVILA~\cite{longvila};
LongVA~\cite{longva}; Qwen2.5-VL~\cite{qwen25vl};
LLaVA-NeXT-Video~\cite{llavanextvideo};
LLaVA-OneVision~\cite{llavaonevision}); Qwen2.5-VL-3B is our own evaluation
under the official protocol. Rank is each method's overall standing by Avg. Best per subtask column in \textbf{bold}; second best
\underline{underlined}.}
\label{tab:vsi-realistic}
\end{table*}

\subsection{GraFT versus Prior Work}
\label{sec:exp-real}

In the last experiment, we test whether GraFT can match models trained for spatial reasoning.
We compare head-to-head with Struct2D on the subsets it reports
(Table~\ref{tab:struct2d}), and on the complete VSI-Bench
under \emph{realistic perception} (Table~\ref{tab:vsi-realistic}).

Against Struct2D, GraFT leads in all three settings Struct2D reports: GPT-o3 under both ground-truth and realistic perception, and Qwen2.5-VL-7B under realistic perception. This holds even though Struct2D is fine-tuned on Struct2D-Set while GraFT does no training. Appendix~\ref{sec:struct2d-app} gives the per-subtask breakdown.

On the complete VSI-Bench, GraFT reaches up to \textbf{51.4} on the official
average, ahead of every general-purpose open-source baseline and of
several prominent fine-tuned spatial models, leading VG LLM-8B, Spatial-MLLM,
OCR, SpaceR, and ViLaSR by $+0.7$ to $+6.0$. The lift is also consistent
across backbones ($+18.8$, $+17.0$, and $+13.9$ over the Qwen2.5-VL-3B,
Qwen2.5-VL-7B, and InternVL2-8B baselines), and with GraFT a 3B model ($47.5$)
surpasses the strongest 72B general-purpose baseline ($40.9$). Additional single-module ablations in Appendix~\ref{sec:single} support the claim that this gain comes from matching each task to the right module rather than any single capability. GraFT thus matches or surpasses models trained specifically for spatial reasoning, all without training of its own.

\section{Conclusion}

We introduced GraFT, a training-free framework that gives a frozen MLLM 3D
spatial reasoning by operating over a single 3DSG. From this 3DSG, GraFT
derives three forms in which the scene's 3D structure reaches the backbone,
each supplying the evidence its task demands. Per-module studies validate each
capability on its target tasks, and GraFT lifts three open-source backbones by
37\% to 65\% on VSI-Bench, reaching 51.4 and surpassing
several models fine-tuned for spatial reasoning; even a 3B backbone overtakes
the strongest 72B baseline.

A frozen MLLM paired with a compact 3DSG is already a strong spatial reasoner. Because the graph is easy to maintain and shared by
every capability, new modules can extend GraFT to further spatial tasks.

%%%%%%%%%%%%%%%%%%%%%%%%%%%%%%%%%%%%%%%%%%%%%%%%%%%%%%%%%%%%%%%%%%%%%%%%%%%%%%%%
%  Acknowledgments: leave commented out for the anonymous ARR submission.
%%%%%%%%%%%%%%%%%%%%%%%%%%%%%%%%%%%%%%%%%%%%%%%%%%%%%%%%%%%%%%%%%%%%%%%%%%%%%%%%
% \section*{Acknowledgments}

\bibliography{references}

\begin{thebibliography}{56}
\providecommand{\natexlab}[1]{#1}

\bibitem[{Armeni et~al.(2019)Armeni, He, Gwak, Zamir, Fischer, Malik, and
  Savarese}]{armeni3dsg}
Armeni, I.; He, Z.-Y.; Gwak, J.; Zamir, A.~R.; Fischer, M.; Malik, J.; and
  Savarese, S. 2019.
\newblock {3D} Scene Graph: A Structure for Unified Semantics, {3D} Space, and
  Camera.
\newblock In \emph{Proceedings of the IEEE/CVF International Conference on
  Computer Vision (ICCV)}.

\bibitem[{Azuma et~al.(2022)Azuma, Miyanishi, Kurita, and Kawanabe}]{scanqa}
Azuma, D.; Miyanishi, T.; Kurita, S.; and Kawanabe, M. 2022.
\newblock {ScanQA}: {3D} Question Answering for Spatial Scene Understanding.
\newblock In \emph{Proceedings of the IEEE/CVF Conference on Computer Vision
  and Pattern Recognition (CVPR)}.

\bibitem[{Bai et~al.(2025)Bai, Chen, Liu, Wang et~al.}]{qwen25vl}
Bai, S.; Chen, K.; Liu, X.; Wang, J.; et~al. 2025.
\newblock {Qwen2.5-VL} Technical Report.
\newblock arXiv:2502.13923.

\bibitem[{Baruch et~al.(2021)Baruch, Chen, Dehghan, Dimry, Feigin, Fu, Gebauer,
  Joffe, Kurz, Schwartz, and Shulman}]{arkitscenes}
Baruch, G.; Chen, Z.; Dehghan, A.; Dimry, T.; Feigin, Y.; Fu, P.; Gebauer, T.;
  Joffe, B.; Kurz, D.; Schwartz, A.; and Shulman, E. 2021.
\newblock {ARKitScenes}: A Diverse Real-World Dataset for {3D} Indoor Scene
  Understanding Using Mobile {RGB-D} Data.
\newblock In \emph{NeurIPS Datasets and Benchmarks Track}.

\bibitem[{Chen et~al.(2024{\natexlab{a}})Chen, Xu, Kirmani, Ichter, Driess,
  Florence, Sadigh, Guibas, and Xia}]{spatialvlm}
Chen, B.; Xu, Z.; Kirmani, S.; Ichter, B.; Driess, D.; Florence, P.; Sadigh,
  D.; Guibas, L.; and Xia, F. 2024{\natexlab{a}}.
\newblock {SpatialVLM}: Endowing Vision-Language Models with Spatial Reasoning
  Capabilities.
\newblock In \emph{Proceedings of the IEEE/CVF Conference on Computer Vision
  and Pattern Recognition (CVPR)}.

\bibitem[{Chen et~al.(2026)Chen, Uy, Song, Ladhak, Murali, Qu, Birchfield,
  Blukis, and Tremblay}]{spacetools}
Chen, S.; Uy, M.~A.; Song, C.~H.; Ladhak, F.; Murali, A.; Qu, Q.; Birchfield,
  S.; Blukis, V.; and Tremblay, J. 2026.
\newblock {SpaceTools}: Tool-Augmented Spatial Reasoning via Double Interactive
  {RL}.
\newblock In \emph{Proceedings of the IEEE/CVF Conference on Computer Vision
  and Pattern Recognition (CVPR)}.

\bibitem[{Chen et~al.(2025)Chen, Xue, Li, Hu et~al.}]{longvila}
Chen, Y.; Xue, F.; Li, D.; Hu, Q.; et~al. 2025.
\newblock {LongVILA}: Scaling Long-Context Visual Language Models for Long
  Videos.
\newblock In \emph{International Conference on Learning Representations
  (ICLR)}.

\bibitem[{Chen et~al.(2024{\natexlab{b}})Chen, Wang, Tian, Ye, Gao, Cui, Tong,
  Hu, Luo, Ma et~al.}]{internvl2}
Chen, Z.; Wang, W.; Tian, H.; Ye, S.; Gao, Z.; Cui, E.; Tong, W.; Hu, K.; Luo,
  J.; Ma, Z.; et~al. 2024{\natexlab{b}}.
\newblock How Far Are We to {GPT-4V}? Closing the Gap to Commercial Multimodal
  Models with Open-Source Suites.
\newblock \emph{Science China Information Sciences}, 67: 220101.

\bibitem[{Cheng et~al.(2024)Cheng, Yin, Fu, Guo, Yang, Kautz, Wang, and
  Liu}]{spatialrgpt}
Cheng, A.-C.; Yin, H.; Fu, Y.; Guo, Q.; Yang, R.; Kautz, J.; Wang, X.; and Liu,
  S. 2024.
\newblock {SpatialRGPT}: Grounded Spatial Reasoning in Vision-Language Models.
\newblock In \emph{Advances in Neural Information Processing Systems}.

\bibitem[{Dai et~al.(2017{\natexlab{a}})Dai, Chang, Savva, Halber, Funkhouser,
  and Nie{\ss}ner}]{scannet}
Dai, A.; Chang, A.~X.; Savva, M.; Halber, M.; Funkhouser, T.; and Nie{\ss}ner,
  M. 2017{\natexlab{a}}.
\newblock {ScanNet}: Richly-Annotated {3D} Reconstructions of Indoor Scenes.
\newblock In \emph{Proceedings of the IEEE Conference on Computer Vision and
  Pattern Recognition (CVPR)}.

\bibitem[{Dai et~al.(2017{\natexlab{b}})Dai, Nie{\ss}ner, Zollh{\"o}fer, Izadi,
  and Theobalt}]{bundlefusion}
Dai, A.; Nie{\ss}ner, M.; Zollh{\"o}fer, M.; Izadi, S.; and Theobalt, C.
  2017{\natexlab{b}}.
\newblock {BundleFusion}: Real-time Globally Consistent {3D} Reconstruction
  Using On-the-fly Surface Reintegration.
\newblock \emph{ACM Transactions on Graphics}.

\bibitem[{Daxberger et~al.(2025)Daxberger, Wenzel, Griffiths, Gang, Lazarow,
  Kohavi, Kang, Eichner, Yang, Dehghan, and Grasch}]{mmspatial}
Daxberger, E.; Wenzel, N.; Griffiths, D.; Gang, H.; Lazarow, J.; Kohavi, G.;
  Kang, K.; Eichner, M.; Yang, Y.; Dehghan, A.; and Grasch, P. 2025.
\newblock {MM-Spatial}: Exploring {3D} Spatial Understanding in Multimodal
  {LLMs}.
\newblock In \emph{Proceedings of the IEEE/CVF International Conference on
  Computer Vision (ICCV)}.

\bibitem[{Fan et~al.(2026)Fan, Zhang, Li, Zhang, Chen, Hu, Wang, Qu, Zhou,
  Wang, Yan, Xu, Theiss, Chen, Li, Tu, Wang, and Ranjan}]{vlm3r}
Fan, Z.; Zhang, J.; Li, R.; Zhang, J.; Chen, R.; Hu, H.; Wang, K.; Qu, H.;
  Zhou, S.; Wang, D.; Yan, Z.; Xu, H.; Theiss, J.; Chen, T.; Li, J.; Tu, Z.;
  Wang, Z.; and Ranjan, R. 2026.
\newblock {VLM-3R}: Vision-Language Models Augmented with Instruction-Aligned
  {3D} Reconstruction.
\newblock In \emph{IEEE/CVF Conference on Computer Vision and Pattern
  Recognition (CVPR)}.

\bibitem[{{Gemini Team}(2024)}]{gemini15}
{Gemini Team}. 2024.
\newblock Gemini 1.5: Unlocking Multimodal Understanding Across Millions of
  Tokens of Context.
\newblock arXiv:2403.05530.

\bibitem[{Gu et~al.(2026)Gu, Zhang, Wei, Chen, Li, and Song}]{spacemindpp}
Gu, B.; Zhang, Z.; Wei, Z.; Chen, Z.; Li, L.; and Song, Z. 2026.
\newblock {SpaceMind++}: Toward Allocentric Cognitive Maps for Spatially
  Grounded Video {MLLMs}.
\newblock arXiv:2605.09449.

\bibitem[{Gu et~al.(2024)Gu, Kuwajerwala, Morin, Jatavallabhula, Sen, Agarwal,
  Rivera, Paul, Ellis, Chellappa, Gan, de~Melo, Tenenbaum, Torralba, Shkurti,
  and Paull}]{conceptgraphs}
Gu, Q.; Kuwajerwala, A.; Morin, S.; Jatavallabhula, K.~M.; Sen, B.; Agarwal,
  A.; Rivera, C.; Paul, W.; Ellis, K.; Chellappa, R.; Gan, C.; de~Melo, C.~M.;
  Tenenbaum, J.~B.; Torralba, A.; Shkurti, F.; and Paull, L. 2024.
\newblock {ConceptGraphs}: Open-Vocabulary {3D} Scene Graphs for Perception and
  Planning.
\newblock In \emph{IEEE International Conference on Robotics and Automation
  (ICRA)}.

\bibitem[{Han et~al.(2025)Han, Zhou, Rong, An, Wang, Wang, Chi, Sheng, and
  Zhang}]{tiger}
Han, Y.; Zhou, E.; Rong, S.; An, J.; Wang, P.; Wang, Z.; Chi, C.; Sheng, L.;
  and Zhang, S. 2025.
\newblock {TIGeR}: Tool-Integrated Geometric Reasoning in Vision-Language
  Models for Robotics.
\newblock arXiv:2510.07181.

\bibitem[{Honerkamp et~al.(2024)Honerkamp, B{\"u}chner, Despinoy, Welschehold,
  and Valada}]{momallm}
Honerkamp, D.; B{\"u}chner, M.; Despinoy, F.; Welschehold, T.; and Valada, A.
  2024.
\newblock Language-Grounded Dynamic Scene Graphs for Interactive Object Search
  with Mobile Manipulation.
\newblock \emph{IEEE Robotics and Automation Letters}.

\bibitem[{Hong et~al.(2023)Hong, Zhen, Chen, Zheng, Du, Chen, and
  Gan}]{hong3dllm}
Hong, Y.; Zhen, H.; Chen, P.; Zheng, S.; Du, Y.; Chen, Z.; and Gan, C. 2023.
\newblock {3D-LLM}: Injecting the {3D} World into Large Language Models.
\newblock In \emph{Advances in Neural Information Processing Systems}.

\bibitem[{Huang et~al.(2024)Huang, Chen, Wang, Huang, Xu, Wang, Liu, Cheng,
  Zhao, Pang, and Zhao}]{chatscene}
Huang, H.; Chen, Y.; Wang, Z.; Huang, R.; Xu, R.; Wang, T.; Liu, L.; Cheng, X.;
  Zhao, Y.; Pang, J.; and Zhao, Z. 2024.
\newblock Chat-Scene: Bridging {3D} Scene and Large Language Models with Object
  Identifiers.
\newblock In \emph{Advances in Neural Information Processing Systems}.

\bibitem[{Huang et~al.(2026)Huang, Zhang, Huang, Yang, and Tang}]{consispace}
Huang, T.; Zhang, Z.; Huang, W.; Yang, J.; and Tang, H. 2026.
\newblock {ConsiSpace}: Learning Geometric Consistency Matters for Video
  Spatial Reasoning.
\newblock arXiv:2607.17599.

\bibitem[{Hughes, Chang, and Carlone(2022)}]{hydra}
Hughes, N.; Chang, Y.; and Carlone, L. 2022.
\newblock Hydra: A Real-time Spatial Perception System for {3D} Scene Graph
  Construction and Optimization.
\newblock In \emph{Proceedings of Robotics: Science and Systems (RSS)}.

\bibitem[{Kolodiazhnyi et~al.(2025)Kolodiazhnyi, Vorontsova, Skripkin,
  Rukhovich, and Konushin}]{unidet3d}
Kolodiazhnyi, M.; Vorontsova, A.; Skripkin, M.; Rukhovich, D.; and Konushin, A.
  2025.
\newblock {UniDet3D}: Multi-dataset Indoor 3D Object Detection.
\newblock In \emph{Proceedings of the AAAI Conference on Artificial
  Intelligence}.

\bibitem[{Li et~al.(2025{\natexlab{a}})Li, Zhang, Guo, Zhang
  et~al.}]{llavaonevision}
Li, B.; Zhang, Y.; Guo, D.; Zhang, R.; et~al. 2025{\natexlab{a}}.
\newblock {LLaVA-OneVision}: Easy Visual Task Transfer.
\newblock \emph{Transactions on Machine Learning Research (TMLR)}.

\bibitem[{Li et~al.(2026{\natexlab{a}})Li, Li, Wang, Yan, Wu, Zhang, Shen, Lu,
  Xiao, and Zhuang}]{spatialladder}
Li, H.; Li, D.; Wang, Z.; Yan, Y.; Wu, H.; Zhang, W.; Shen, Y.; Lu, W.; Xiao,
  J.; and Zhuang, Y. 2026{\natexlab{a}}.
\newblock {SpatialLadder}: Progressive Training for Spatial Reasoning in
  Vision-Language Models.
\newblock In \emph{International Conference on Learning Representations
  (ICLR)}.

\bibitem[{Li et~al.(2025{\natexlab{b}})Li, Song, Li, Guo, Yao, Xu, Liu, and
  Xiong}]{seetrek}
Li, P.; Song, P.; Li, W.; Guo, W.; Yao, H.; Xu, Y.; Liu, D.; and Xiong, H.
  2025{\natexlab{b}}.
\newblock {See\&Trek}: Training-Free Spatial Prompting for Multimodal Large
  Language Model.
\newblock In \emph{Advances in Neural Information Processing Systems}.

\bibitem[{Li et~al.(2026{\natexlab{b}})Li, Mu, Li, Qian, Zhang, Hou, and
  Cheng}]{annorollouts}
Li, Y.; Mu, G.; Li, H.; Qian, S.; Zhang, D.; Hou, Q.; and Cheng, M.-M.
  2026{\natexlab{b}}.
\newblock Annotations as Rollouts: Efficient and Scalable Reinforcement
  Learning for Video {MLLMs}.
\newblock arXiv:2608.20492.

\bibitem[{Liao et~al.(2025)Liao, Xie, Zhang, Kong, Lu, Yang, and
  Deng}]{r1zerovsi}
Liao, Z.; Xie, Q.; Zhang, Y.; Kong, Z.; Lu, H.; Yang, Z.; and Deng, Z. 2025.
\newblock Improved Visual-Spatial Reasoning via {R1-Zero}-Like Training.
\newblock In \emph{ICCV Workshop on Reliable and Interactable World Models}.

\bibitem[{Lin et~al.(2024)Lin, Yin, Ping, Lu, Molchanov, Tao, Mao, Kautz,
  Shoeybi, and Han}]{vila}
Lin, J.; Yin, H.; Ping, W.; Lu, Y.; Molchanov, P.; Tao, A.; Mao, H.; Kautz, J.;
  Shoeybi, M.; and Han, S. 2024.
\newblock {VILA}: On Pre-Training for Visual Language Models.
\newblock In \emph{Proceedings of the IEEE/CVF Conference on Computer Vision
  and Pattern Recognition (CVPR)}.

\bibitem[{Liu et~al.(2025)Liu, Xue, Wang, Yin, Yang, and Gao}]{spatialsurvey}
Liu, W.; Xue, Q.; Wang, H.; Yin, X.; Yang, B.; and Gao, W. 2025.
\newblock Spatial Reasoning in Multimodal Large Language Models: A Survey of
  Tasks, Benchmarks and Methods.
\newblock arXiv:2511.15722.

\bibitem[{{OpenAI}(2024)}]{gpt4o}
{OpenAI}. 2024.
\newblock {GPT-4o} System Card.
\newblock arXiv:2410.21276.

\bibitem[{Ouyang et~al.(2025)Ouyang, Liu, Wu, Liu, Zhou, Zhou, Meng, and
  Sun}]{spacer}
Ouyang, K.; Liu, Y.; Wu, H.; Liu, Y.; Zhou, H.; Zhou, J.; Meng, F.; and Sun, X.
  2025.
\newblock {SpaceR}: Reinforcing {MLLMs} in Video Spatial Reasoning.
\newblock arXiv:2504.01805.

\bibitem[{Qi et~al.(2026)Qi, Zhang, Fang, Wang, and Zhao}]{gpt4scene}
Qi, Z.; Zhang, Z.; Fang, Y.; Wang, J.; and Zhao, H. 2026.
\newblock {GPT4Scene}: Understand {3D} Scenes from Videos with Vision-Language
  Models.
\newblock In \emph{International Conference on Learning Representations
  (ICLR)}.

\bibitem[{Rotondi et~al.(2026)Rotondi, Argenziano, Koch, Hughes, Buechner,
  Wald, Schmid, Nardi, Valada, Paull, Tombari, Carlone, and Arras}]{sg3dsurvey}
Rotondi, D.; Argenziano, F.; Koch, S.; Hughes, N.; Buechner, M.; Wald, J.;
  Schmid, L.~R.; Nardi, D.; Valada, A.; Paull, L.; Tombari, F.; Carlone, L.;
  and Arras, K.~O. 2026.
\newblock {3D} Scene Graphs: Open Challenges and Future Directions.
\newblock arXiv:2606.19383.

\bibitem[{Saxena et~al.(2025)Saxena, Buchanan, Paxton, Liu, Chen, Vaskevicius,
  Palmieri, Francis, and Kroemer}]{grapheqa}
Saxena, S.; Buchanan, B.; Paxton, C.; Liu, P.; Chen, B.; Vaskevicius, N.;
  Palmieri, L.; Francis, J.; and Kroemer, O. 2025.
\newblock {GraphEQA}: Using {3D} Semantic Scene Graphs for Real-time Embodied
  Question Answering.
\newblock In \emph{Conference on Robot Learning (CoRL)}.

\bibitem[{Schult et~al.(2023)Schult, Engelmann, Hermans, Litany, Tang, and
  Leibe}]{mask3d}
Schult, J.; Engelmann, F.; Hermans, A.; Litany, O.; Tang, S.; and Leibe, B.
  2023.
\newblock {Mask3D}: Mask Transformer for 3D Semantic Instance Segmentation.
\newblock In \emph{IEEE International Conference on Robotics and Automation
  (ICRA)}.

\bibitem[{Tang et~al.(2026)Tang, Cao, Liu, Liang, Li, Li, and Liang}]{ocr}
Tang, H.; Cao, M.; Liu, R.; Liang, X.; Li, L.; Li, G.; and Liang, X. 2026.
\newblock Video Spatial Reasoning with Object-Centric {3D} Rollout.
\newblock In \emph{Proceedings of the AAAI Conference on Artificial
  Intelligence (AAAI)}, 9395--9403.

\bibitem[{Wang(2026)}]{wang2026astra}
Wang, Y. 2026.
\newblock How Well Does {Astra} Understand Real-World Space?
\newblock \url{https://www.yipeng.dev/blog/astra-spatial-intelligence}.
\newblock Blog post, 22 September 2026.

\bibitem[{Wang et~al.(2025)Wang, Huang, Zhao, Zhang, Jin, and Zhao}]{chat3d}
Wang, Z.; Huang, H.; Zhao, Y.; Zhang, Z.; Jin, T.; and Zhao, Z. 2025.
\newblock Data-Efficiently Learn Large Language Model for Universal {3D} Scene
  Perception.
\newblock In \emph{Findings of the Association for Computational Linguistics:
  NAACL}, 313--333.

\bibitem[{Werby et~al.(2024)Werby, Huang, B{\"u}chner, Valada, and
  Burgard}]{hovsg}
Werby, A.; Huang, C.; B{\"u}chner, M.; Valada, A.; and Burgard, W. 2024.
\newblock Hierarchical Open-Vocabulary {3D} Scene Graphs for Language-Grounded
  Robot Navigation.
\newblock In \emph{Robotics: Science and Systems (RSS)}.

\bibitem[{Wu et~al.(2025{\natexlab{a}})Wu, Liu, Hung, and Duan}]{spatialmllm}
Wu, D.; Liu, F.; Hung, Y.-H.; and Duan, Y. 2025{\natexlab{a}}.
\newblock Spatial-{MLLM}: Boosting {MLLM} Capabilities in Visual-Based Spatial
  Intelligence.
\newblock In \emph{Advances in Neural Information Processing Systems}.

\bibitem[{Wu et~al.(2026)Wu, Huang, Chen, Zhang, Wang, and Xie}]{spatialscore}
Wu, H.; Huang, X.; Chen, Y.; Zhang, Y.; Wang, Y.; and Xie, W. 2026.
\newblock {SpatialScore}: Towards Comprehensive Evaluation for Spatial
  Intelligence.
\newblock In \emph{Proceedings of the IEEE/CVF Conference on Computer Vision
  and Pattern Recognition (CVPR)}.

\bibitem[{Wu et~al.(2025{\natexlab{b}})Wu, Guan, Feng, Liu, Wu, Wang, Wu, and
  Tan}]{vilasr}
Wu, J.; Guan, J.; Feng, K.; Liu, Q.; Wu, S.; Wang, L.; Wu, W.; and Tan, T.
  2025{\natexlab{b}}.
\newblock Reinforcing Spatial Reasoning in Vision-Language Models with
  Interwoven Thinking and Visual Drawing.
\newblock In \emph{Advances in Neural Information Processing Systems}.

\bibitem[{Xu et~al.(2024)Xu, Wang, Wang, Chen, Pang, and Lin}]{pointllm}
Xu, R.; Wang, X.; Wang, T.; Chen, Y.; Pang, J.; and Lin, D. 2024.
\newblock {PointLLM}: Empowering Large Language Models to Understand Point
  Clouds.
\newblock In \emph{Proceedings of the European Conference on Computer Vision
  (ECCV)}.

\bibitem[{Yang et~al.(2025)Yang, Yang, Gupta, Han, Li, and Xie}]{vsibench}
Yang, J.; Yang, S.; Gupta, A.~W.; Han, R.; Li, F.-F.; and Xie, S. 2025.
\newblock Thinking in Space: How Multimodal Large Language Models See,
  Remember, and Recall Spaces.
\newblock In \emph{Proceedings of the IEEE/CVF Conference on Computer Vision
  and Pattern Recognition (CVPR)}.

\bibitem[{Yang et~al.(2023)Yang, Zhang, Li, Zou, Li, and Gao}]{som}
Yang, J.; Zhang, H.; Li, F.; Zou, X.; Li, C.; and Gao, J. 2023.
\newblock Set-of-Mark Prompting Unleashes Extraordinary Visual Grounding in
  {GPT-4V}.
\newblock arXiv:2310.11441.

\bibitem[{Yang et~al.(2026)Yang, Zhao, Pan, Yang, Zhang, Kang, Xu, Li, and
  Xie}]{cambrianp}
Yang, J.; Zhao, Z.; Pan, X.; Yang, S.; Zhang, J.; Kang, B.; Xu, H.; Li, S.-W.;
  and Xie, S. 2026.
\newblock {Cambrian-P}: Pose-Grounded Video Understanding.
\newblock arXiv:2605.22819.

\bibitem[{Yao et~al.(2025)Yao, Yu, Zhang, Wang, Cui, Zhu, Cai, Li, Zhao, He
  et~al.}]{minicpmv}
Yao, Y.; Yu, T.; Zhang, A.; Wang, C.; Cui, J.; Zhu, H.; Cai, T.; Li, H.; Zhao,
  W.; He, Z.; et~al. 2025.
\newblock Efficient {GPT-4V} Level Multimodal Large Language Model for
  Deployment on Edge Devices.
\newblock \emph{Nature Communications}, 16: 5509.

\bibitem[{Yeshwanth et~al.(2023)Yeshwanth, Liu, Nie{\ss}ner, and
  Dai}]{scannetpp}
Yeshwanth, C.; Liu, Y.-C.; Nie{\ss}ner, M.; and Dai, A. 2023.
\newblock {ScanNet++}: A High-Fidelity Dataset of {3D} Indoor Scenes.
\newblock In \emph{Proceedings of the IEEE/CVF International Conference on
  Computer Vision (ICCV)}.

\bibitem[{Yin et~al.(2024)Yin, Xu, Wu, Zhou, and Lu}]{sgnav}
Yin, H.; Xu, X.; Wu, Z.; Zhou, J.; and Lu, J. 2024.
\newblock {SG-Nav}: Online {3D} Scene Graph Prompting for {LLM}-based Zero-shot
  Object Navigation.
\newblock In \emph{Advances in Neural Information Processing Systems
  (NeurIPS)}.

\bibitem[{Yuan, Kumar, and Wang(2026)}]{gr3d}
Yuan, J.; Kumar, G.; and Wang, B. 2026.
\newblock Boosting {MLLM} Spatial Reasoning with Geometrically Referenced {3D}
  Scene Representations.
\newblock arXiv:2603.08592.

\bibitem[{Zhang et~al.(2025{\natexlab{a}})Zhang, Hao, Xu, Zhang, Zhang, Wang,
  Zhang, Wang, Zhang, and Xu}]{mapnav}
Zhang, L.; Hao, X.; Xu, Q.; Zhang, Q.; Zhang, X.; Wang, P.; Zhang, J.; Wang,
  Z.; Zhang, S.; and Xu, R. 2025{\natexlab{a}}.
\newblock {MapNav}: A Novel Memory Representation via Annotated Semantic Maps
  for Vision-and-Language Navigation.
\newblock In \emph{Proceedings of the 63rd Annual Meeting of the Association
  for Computational Linguistics (ACL)}.

\bibitem[{Zhang et~al.(2025{\natexlab{b}})Zhang, Zhang, Li, Zeng
  et~al.}]{longva}
Zhang, P.; Zhang, K.; Li, B.; Zeng, G.; et~al. 2025{\natexlab{b}}.
\newblock Long Context Transfer from Language to Vision.
\newblock \emph{Transactions on Machine Learning Research (TMLR)}.

\bibitem[{Zhang et~al.(2024)Zhang, Li, Liu, Lee, Gui, Fu, Feng, Liu, and
  Li}]{llavanextvideo}
Zhang, Y.; Li, B.; Liu, H.; Lee, Y.~J.; Gui, L.; Fu, D.; Feng, J.; Liu, Z.; and
  Li, C. 2024.
\newblock {LLaVA-NeXT}: A Strong Zero-Shot Video Understanding Model.
\newblock \url{https://llava-vl.github.io/blog/2024-04-30-llava-next-video/}.

\bibitem[{Zheng et~al.(2025)Zheng, Huang, Li, and Wang}]{vgllm}
Zheng, D.; Huang, S.; Li, Y.; and Wang, L. 2025.
\newblock Learning from Videos for {3D} World: Enhancing {MLLMs} with {3D}
  Vision Geometry Priors.
\newblock In \emph{Advances in Neural Information Processing Systems}.

\bibitem[{Zhu et~al.(2025)Zhu, Wang, Xie, Gu, Ding, Yang, and Jiang}]{struct2d}
Zhu, F.; Wang, H.; Xie, Y.; Gu, J.; Ding, T.; Yang, J.; and Jiang, H. 2025.
\newblock Struct2D: A Perception-Guided Framework for Spatial Reasoning in
  {MLLMs}.
\newblock In \emph{Advances in Neural Information Processing Systems}.

\end{thebibliography}

% Start the appendix on a fresh page, the way the AAAI submission shipped it as
% a separate Supplementary Material document. \clearpage is forbidden in an
% AAAI *submission*; this is an arXiv preprint, so that rule does not apply --
% do not copy this line back into ../AAAI/.
\clearpage

\appendix
\makeatletter
\@addtoreset{table}{section}
\@addtoreset{figure}{section}
\makeatother
\setcounter{table}{0}
\setcounter{figure}{0}
\renewcommand{\thetable}{\thesection\arabic{table}}
\renewcommand{\thefigure}{\thesection\arabic{figure}}
\renewcommand{\theequation}{\thesection\arabic{equation}}

\section{Implementation Details}
\label{sec:impl}

\subsection{3DSG Construction}
Under ground-truth perception we build the graph from the dataset annotations of
ARKitScenes, ScanNet, and ScanNet++. Under realistic perception we follow the Struct2D setup: the scene is reconstructed offline and objects are detected with UniDet3D and Mask3D; and converted into the same graph structure as the ground-truth 3DSG, which serves as the interface for our reasoning modules. Per-frame camera poses and intrinsics are kept with the graph for
the egocentric module.

\subsection{Router}
\label{sec:router-prompt}

Section~\ref{sec:router} routes each task with one plaintext pass through the
backbone. The system prompt for that pass is reproduced below (Router prompt).
It names the module labels, explains what each covers; the decode is constrained to those labels by a JSON schema, so the
returned value is always one of them. The three labels map onto the three modules: \emph{tools} to the symbolic module,
\emph{bev} to the BEV module, \emph{ego} to the egocentric module.

\begin{promptbox}{Router prompt}
You are a router. Pick exactly one of three outputs. Return JSON {"branch": one of [tools|bev|ego]}.

Branches:
- tools: exact numeric values from geometry (counts, distances, dimensions, room size and so on).
- bev: tasks that need an egocentric-to-allocentric conversion, where the observer stands at an anchor object and the answer must be given in that observer's frame. NEVER for proximity / closeness / distance questions (those go to tools).
- ego: fine-grained appearance (color, material, texture, shape of one object).

\end{promptbox}

\subsection{Symbolic Tools}
\label{sec:tools-app}

The symbolic module of Section~\ref{sec:tools} exposes five tools, each reading
directly from the object boxes of the 3DSG:
\begin{itemize}
  \item a \emph{count} tool that lists each object class with its instance count;
  \item a \emph{size} tool that returns an object's width, depth, and height as
    the extents of its box;
  \item a \emph{distance} tool that measures the surface-to-surface distance
    between the boxes of two object classes;
  \item a \emph{proximity} tool that ranks several candidate classes by
    their closest distance to an anchor object, resolving which is nearest or
    farthest;
  \item a \emph{room} tool that returns the room's width, depth, height, and
    floor area.
\end{itemize}

\subsection{BEV for Relative Direction}
\label{sec:bev-rel}

For a relative-direction question, GraFT renders a single query-conditioned,
heading-aligned bird's-eye view (BEV) of the 3DSG and reads the answer from it.
Algorithm~\ref{alg:bev-rel} details the rendering. \textsc{RelevantClasses} returns the object classes a question refers to, and \textsc{AnchorFacing} the anchor object and the object it faces. Both are produced by a single constrained-decode call that picks them from the classes present in the scene graph and attaches an instance index to each. The BEV is an orthographic
top-down projection onto the world $XY$ plane, a fixed $640\times480$ image; the
scene is rotated so the observer's facing direction points up ($+Y=$ forward),
each object is labeled with a class number and instance suffix, and a dashed arrow points from
the anchor to the queried target. A parallel textual key lists each drawn
instance's observer-frame coordinates so the backbone can cross-reference the
markers.

\begin{algorithm}[!ht]
\caption{BEV rendering for relative direction}
\label{alg:bev-rel}
\begin{algorithmic}[1]
\Require 3DSG $\mathcal{G}$, question $q$
\State $\mathcal{C} \gets \textsc{RelevantClasses}(q)$ \Comment{resolved against $\mathcal{G}$'s classes, with instance indices}
\State \textbf{if} $\mathcal{C} = \emptyset$ \textbf{then} $\mathcal{C} \gets \mathrm{classes}(\mathcal{G})$ \Comment{fall back to the whole scene}
\State $(a, f) \gets \textsc{AnchorFacing}(q)$ \Comment{anchor object $a$, faced object $f$}
\State $\theta \gets \tfrac{\pi}{2} - \operatorname{atan2}\big((\mathbf{c}_f{-}\mathbf{c}_a)_y,\, (\mathbf{c}_f{-}\mathbf{c}_a)_x\big)$ \Comment{rotate so $\mathbf{c}_a\!\to\!\mathbf{c}_f$ points up}
\State initialize blank canvas $I$
\For{$o \in \mathcal{G}$ with $\mathrm{class}(o)\in\mathcal{C}$}
  \State $P \gets$ footprint of $o$, rotated by $\theta$ about $\mathbf{c}_a$
  \State draw filled $P$ on $I$ with label
\EndFor
\State draw dashed arrow $a \to$ target
\State $\mathcal{K} \gets \{(\mathrm{label}(o), x_o, y_o)\}$ in the observer frame \Comment{$+x{=}$right, $+y{=}$forward}
\Ensure BEV image $I$ and object key $\mathcal{K}$
\end{algorithmic}
\end{algorithm}

The rendered BEV and its key are sent to the backbone as the following instruction (Relative direction prompt).

\begin{promptbox}{Relative direction prompt}
The image is a top-down (BEV) view of the room with you at the anchor. The BEV is heading-aligned, so your facing direction points UP: image top = FRONT, bottom = BACK, left = LEFT, right = RIGHT. A dashed dark-blue arrow goes from the observer's position to the TARGET named in the question. Classify which quadrant the arrow enters and map it to the listed options; never pick a direction that is not listed.
Object positions in your observer reference frame:
    1 stove: [+0.0, +0.0]
    2 sofa: [+0.0, +4.2]
    3 tv: [-2.2, +2.8]
\end{promptbox}

\subsection{BEV for Route Planning}
\label{sec:bev-route}

Route planning uses the same BEV rendering with a route-specific layout
(Algorithm~\ref{alg:bev-route}). \textsc{ParseRoute} returns the route's landmarks, the start, the
initial facing landmark, an ordered list of waypoints, and the final target,
each as a scene class with an instance index, from the same kind of
constrained-decode call; it chooses the instance combination that yields a
coherent, backtrack-free route. A landmark the scene graph does not contain is
returned with an estimated position instead, and is drawn and named as an
estimate. All objects are drawn and numbered as marks, and the image
is paired with a legend and an ordered step list; the backbone decides each turn
from the step list and the image. Marks are numbered in the order the route
visits them, and each forward step names the next mark together with its
coordinate, so the backbone receives the route geometry
numerically as well as through the rendered marks.

The initial heading comes from the start and facing landmarks. When they are
distinct objects, it is the floor-plane direction $\mathbf{c}_f-\mathbf{c}_s$ from the start
centre to the facing centre. When both are the \emph{same} object (``standing at the sofa,
facing the sofa''), that is undefined, so the heading is the negated floor-plane
surface normal of the start object $-\mathbf{n}^{xy}_s$.
The scene is rotated about the start centre $\mathbf{c}_s$ so this heading lands on $+Y$.

\begin{algorithm}[!ht]
\caption{BEV rendering for route planning}
\label{alg:bev-route}
\begin{algorithmic}[1]
\Require 3DSG $\mathcal{G}$, question $q$
\State $(s, f, [w_1,\dots,w_m], g) \gets \textsc{ParseRoute}(q, \mathcal{G})$ \Comment{start, facing, waypoints, target}
\State \textbf{if} a landmark $\notin \mathcal{G}$ \textbf{then} estimate its floor position \Comment{absent landmark, never left unresolved}
\State \textbf{if} $f\neq s$ \textbf{then} $\mathbf{h}\gets \mathbf{c}_f-\mathbf{c}_s$ \textbf{else} $\mathbf{h}\gets-\mathbf{n}^{xy}_s$ \Comment{facing: start$\to$facing, or start's inward normal}
\State $\theta \gets$ heading so $\mathbf{h}$ points up ($+Y$)
\State initialize blank canvas $I$
\For{$o \in \mathcal{G}$}
  \State $P \gets$ footprint of $o$, rotated by $\theta$ about $\mathbf{c}_s$
  \State draw filled $P$ on $I$ with a mark label
\EndFor
\State $L \gets$ ordered steps ``\textit{Step $k$: go forward to mark $j$ ($\mathrm{name}$) at $(x,y)$}''
\Ensure BEV image $I$, step list $L$
\end{algorithmic}
\end{algorithm}

The rendered BEV and its step list are sent to the backbone as the following instruction (Route prompt).

\begin{promptbox}[breakable=false]{Route prompt}
You are a robot. The image is a top-down (BEV) view of the room with you at the
anchor. The BEV is heading-aligned, so your INITIAL facing direction points UP:
image top = FRONT, bottom = BACK, left = LEFT, right = RIGHT
(+y = forward, -y = back, +x = right, -x = left).
After each "go forward to X" your actual facing rotates to the direction you
just walked, but the BEV image stays fixed in this initial frame.

Object marks: 1=sofa (anchor), 2=fireplace, 3=stove (target)
Route:
  Step 1: Go forward to mark 2 (fireplace) at (0.00, 2.41).
  Step 2: [turn left / turn right / turn back]
  Step 3: Go forward until mark 3 (stove) at (-5.47, -0.62).
Judge each turn qualitatively from the BEV image: see where the next target
lies relative to your current facing, and pick the matching turn.
\end{promptbox}

\subsection{Egocentric Frame Retrieval}
\label{sec:ego-impl}

For a visual-attribute question, GraFT scores every posed frame for the queried
object and forwards the top-$k$ to the backbone (Algorithm~\ref{alg:ego};
$k{=}2$ on ScanQA). A frame's score multiplies an inverse-distance proximity term, a centrality term ($r$ is the projected object's pixel offset from the image centre, $\rho$ the half-diagonal), and a coverage term (the fraction of the object's eight box corners inside the image), gated by the visibility thresholds of Table~\ref{tab:defaults}.

All thresholds are fixed once and shared across scenes and tasks;
Table~\ref{tab:defaults} lists the values. The depth band and the corner and
viewing-angle thresholds instantiate the visibility gate of this module
(Algorithm~\ref{alg:ego}). The de-duplication window controls the diversity of
the forwarded set: consecutive video frames are nearly identical, so after
ranking, a candidate is skipped whenever it lies within $15$ positions of an already selected frame in the chronologically ordered frame list.

\begin{center}
\small
\setlength{\tabcolsep}{1.4mm}
\begin{tabular}{llc}
\toprule
Parameter & Symbol & Value \\
\midrule
Depth band & $d_{\min}$, $d_{\max}$ & $0.5$\,m, $4$\,m \\
Viewing-angle threshold & $\cos\tau$ & $0.5$ \\
Corners required in frame & $n$ & $\ge 2$ of $8$ \\
Frame budget (ScanQA) & $k$ & $2$ \\
De-duplication window & -- & $15$ frames \\
Visibility threshold & -- & $0$ \\
\bottomrule
\end{tabular}
\captionof{table}{Default parameters of the egocentric module, fixed across
scenes and tasks.}
\label{tab:defaults}
\end{center}

\begin{algorithm}[!ht]
\caption{Egocentric frame retrieval}
\label{alg:ego}
\begin{algorithmic}[1]
\Require 3DSG $\mathcal{G}$, question $q$, posed frames $(\mathrm{pose}_t, K_t)_{t=1}^{N}$, budget $k$
\State $\mathcal{T} \gets$ target objects of $q$ resolved against $\mathcal{G}$ \Comment{by class name}
\For{each frame $t$}
  \State $s_t \gets 0$;\ \ $c_t \gets 0$
  \For{each target $o \in \mathcal{T}$}
    \State project $o$'s centre to pixel $\mathbf{u}$ at depth $d$;\ $n \gets \#$ corners of $o$ inside image \Comment{pinhole $K_t,\mathrm{pose}_t$}
    \If{$\langle\mathbf{z}_t,\mathbf{c}_o{-}\mathbf{p}_t\rangle/d \ge \cos\tau$, $0.5\!\le\!d\!\le\!4$, and $n\!\ge\!2$}
      \State $s_t \gets s_t + \tfrac{1}{\max(d,1)}\cdot\max(0,\,1{-}r/\rho)\cdot\tfrac{n}{8}$ \Comment{prox.\ $\times$ centrality $\times$ coverage}
      \State $c_t \gets c_t + 1$
    \EndIf
  \EndFor
\EndFor
\State rank frames by $(c_t, s_t)$ descending
\State $F \gets$ top-$k$, greedily skipping frames temporally close to a chosen one
\State \textbf{if} $F=\emptyset$ \textbf{then} $F \gets k$ uniformly spaced frames \Comment{fallback}
\Ensure selected egocentric frames $F$
\end{algorithmic}
\end{algorithm}

\paragraph{Appearance order on VSI-Bench.}
For VSI-Bench's appearance-order task, the same per-frame visibility score is
used under a temporal rule instead of ranking: for each queried object, GraFT
scans its frames in chronological order and records the first frame whose visibility score passes the score threshold of Table~\ref{tab:defaults}; the objects are then ordered by
these first-visible times, and that order is the answer.

\section{Additional Experiments}
\label{sec:more}

\subsection{Comparison with Struct2D}
\label{sec:struct2d-app}

Table~\ref{tab:struct2d} gives the per-subtask breakdown behind the comparison
in Section~\ref{sec:exp-real}, on the VSI-Bench subsets that Struct2D reports.
Two of the three settings use a GPT-o3 backbone and compare on six subtasks,
where GraFT gains $+9.3$ average points under \emph{ground-truth perception},
ahead on five of the six, and $+2.7$ under \emph{realistic perception}. The third uses a
Qwen2.5-VL-7B backbone over all seven subtasks Struct2D reports, where GraFT
gains $+6.8$ under \emph{realistic perception}.

\begin{table}[!t]
\centering
\small
\setlength{\tabcolsep}{0.6mm}
\begin{tabular}{lcccccccc}
\toprule
Method & Avg. & \shortstack{Obj.\\ Cnt.} & \shortstack{Abs.\\ Dist.} & \shortstack{Obj.\\ Size} & \shortstack{Room\\ Size} & \shortstack{Rel.\\ Dist.} & \shortstack{Rel.\\ Dir.} & \shortstack{Route\\ Plan} \\
\midrule
\multicolumn{9}{l}{\textbf{GPT-o3}} \\
\rowcolor{grouprow}\multicolumn{9}{l}{\textit{Ground-truth perception}} \\
\quad Struct2D & 83.8 & 93.8 & 90.6 & -- & 47.4 & 96.5 & 94.4 & 80.1 \\
\quad GraFT & \textbf{93.1} & 90.1 & 97.5 & -- & 91.1 & 98.2 & 96.1 & 85.7 \\
\cmidrule(lr){1-9}
\rowcolor{grouprow}\multicolumn{9}{l}{\textit{Realistic perception}} \\
\quad Struct2D & 56.1 & 52.8 & 38.4 & -- & 48.9 & 60.0 & 60.1 & 76.2 \\
\quad GraFT & \textbf{58.8} & 49.5 & 42.7 & -- & 72.3 & 45.5 & 71.6 & 71.4 \\
\midrule
\multicolumn{9}{l}{\textbf{Qwen2.5-VL-7B}} \\
\rowcolor{grouprow}\multicolumn{9}{l}{\textit{Realistic perception}} \\
\quad Struct2D & 43.6 & 47.1 & 35.1 & 57.1 & 48.9 & 35.1 & 45.9 & 35.8 \\
\quad GraFT & \textbf{50.4} & 44.6 & 42.3 & 67.5 & 83.4 & 43.4 & 42.8 & 28.9 \\
\bottomrule
\end{tabular}
\caption{Our GraFT vs.\ Struct2D~\cite{struct2d}. Top: zero-shot GPT-o3 on the six subtasks of the VSI-Bench subset that Struct2D reports on (Obj.\ Size not reported,
marked ``--''), under ground-truth and realistic perception. Bottom: Qwen2.5-VL-7B on the seven \emph{VSI-Bench} subtasks Struct2D reports, under realistic perception. Struct2D is fine-tuned (SFT) while GraFT is training-free. Avg.\
averages each panel's own subtasks; the higher Avg.\ per group is in
\textbf{bold}.}
\label{tab:struct2d}
\end{table}

\begin{table*}[t]
\centering
\small
\setlength{\tabcolsep}{1mm}
\begin{tabular}{lccccccccc}
\toprule
Method & Avg. & \shortstack{Obj.\\ Cnt.} & \shortstack{Abs.\\ Dist.} & \shortstack{Obj.\\ Size} & \shortstack{Room\\ Size} & \shortstack{Rel.\\ Dist.} & \shortstack{Rel.\\ Dir.} & \shortstack{Route\\ Plan} & \shortstack{App.\\ Ord.} \\
\midrule
Tools  & 45.9 & 44.6 & 42.3 & 67.5 & 83.4 & 43.4 & 37.3 & 30.4 & 18.0 \\
BEV    & 29.4 & 45.4 & 20.6 & 33.7 & 27.8 & 29.4 & 42.8 & 28.9 & 6.5 \\
Ego    & 35.4 & 38.2 & 13.8 & 45.5 & 32.8 & 36.9 & 37.0 & 33.0 & 46.8 \\
\bottomrule
\end{tabular}
\caption{Single-module ablation on VSI-Bench (Qwen2.5-VL-7B, realistic
perception). Each row applies a single module to all eight subtasks; the full
GraFT that routes per question scores $50.0$ (Table~\ref{tab:vsi-realistic}). Avg.\ is over
the eight subtasks.}
\label{tab:single-branch}
\end{table*}

\subsection{Single-Module Ablation}
\label{sec:single}

Table~\ref{tab:single-branch} isolates the VSI-Bench capabilities by applying one module to \emph{all} eight subtasks, against the full GraFT (Table~\ref{tab:vsi-realistic}). The results show that each module leads on its target family. Tools dominates the metric subtasks (e.g.\ room size $83.4$ vs.\ $32.8$ and $27.8$), BEV leads relative direction ($42.8$), and Ego leads appearance order ($46.8$), whose temporal ordering only frame retrieval carries. Selecting the right module per question is what lifts the full system ($50.0$) above every single module ($45.9$ / $29.4$ / $35.4$).

\subsection{Egocentric Frame Budget}
\label{sec:kablation}

Table~\ref{tab:k-ablation} ablates the number of retrieved egocentric frames $k$
on ScanQA. The main text reports $k{=}2$ (Table~\ref{tab:ego-scanqa}); here we vary
$k\in\{1,2,3\}$. Scores rise steadily with the frame budget on most metrics,
for instance CIDEr ($69.2\to73.6\to75.9$); $k{=}2$ already recovers most of the gain and
gives the best BLEU-4, motivating it as the default. Notably, even a single
selected frame outperforms uniform 8-frame sampling: the gain comes from the geometric selection
rule, not from the frame count.

\begin{table}[!ht]
\centering
\small
\setlength{\tabcolsep}{0.6mm}
\begin{tabular}{lccccc}
\toprule
Frames & B-1 & B-4 & METEOR & ROUGE & CIDEr \\
\midrule
Uniform sampling & 22.2 & 7.4 & 16.6 & 30.8 & 58.0 \\
\cmidrule(lr){1-6}
Selected top-k, $k{=}1$ & 32.0 & 8.0 & 20.6 & 36.1 & 69.2 \\
Selected top-k, $k{=}2$ & 34.2 & 9.8 & 21.7 & 38.0 & 73.6 \\
Selected top-k, $k{=}3$ & 35.5 & 8.7 & 22.5 & 39.5 & 75.9 \\
\bottomrule
\end{tabular}
\caption{Egocentric frame-budget ablation on ScanQA (Qwen2.5-VL-7B, realistic
perception). \emph{Selected} rows retrieve the top-$k$ geometry-ranked frames;
\emph{Uniform} samples $8$ frames evenly (Table~\ref{tab:ego-scanqa}). $k{=}2$
is the default used in the main text.}
\label{tab:k-ablation}
\end{table}

\subsection{Upper Bound with Ground-Truth Perception}
\label{sec:upper-bound}

\begin{figure}[t]
\centering
\includegraphics[width=\columnwidth]{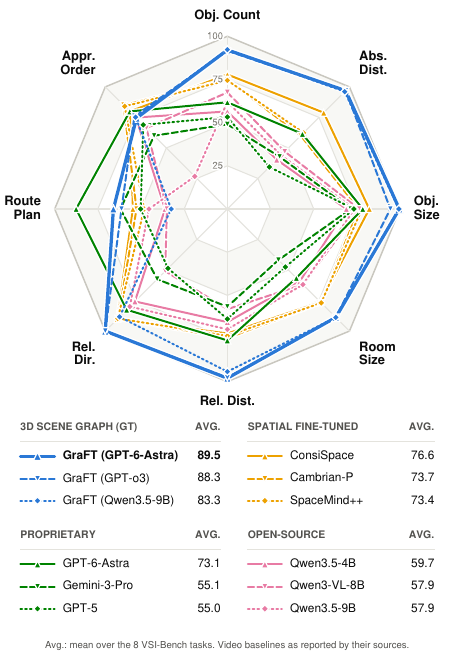}
\caption{GraFT with ground-truth perception against the strongest video-input
models on VSI-Bench: a subset of the rows of Table~\ref{tab:upper-bound}, with
the same numbers.}
\label{fig:upper-bound}
\end{figure}

\begin{table*}[t]
\centering
\small
\setlength{\tabcolsep}{1mm}
% Generated by scripts/o3/plot_vsi_upper_bound.py from scripts/o3/vsi_upper_bound.csv
\begin{tabular}{lccccccccc}
\toprule
Method & Avg. & \shortstack{Obj.\\ Cnt.} & \shortstack{Abs.\\ Dist.} & \shortstack{Obj.\\ Size} & \shortstack{Room\\ Size} & \shortstack{Rel.\\ Dist.} & \shortstack{Rel.\\ Dir.} & \shortstack{Route\\ Plan} & \shortstack{App.\\ Ord.} \\
\midrule
\rowcolor{grouprow}\multicolumn{10}{l}{\textit{GraFT, ground-truth 3DSG (ours)}} \\
GraFT (GPT-6-Astra) & \textbf{89.5} & 92.1 & \textbf{96.9} & \textbf{99.6} & 88.8 & \textbf{98.0} & \textbf{100.0} & \textbf{66.0} & 74.4 \\
GraFT (GPT-o3) & 88.3 & 91.6 & \textbf{96.9} & 94.4 & 88.8 & 97.7 & \textbf{100.0} & 61.3 & 76.1 \\
GraFT (GPT-4.1) & 85.3 & 91.5 & \textbf{96.9} & 94.5 & 88.8 & 95.8 & 98.1 & 41.2 & 75.6 \\
GraFT (Qwen3.5-9B) & 83.3 & \textbf{92.3} & 96.1 & 99.3 & 88.8 & 94.2 & 88.3 & 32.5 & 75.2 \\
GraFT (Qwen3.5-2B) & 79.1 & 86.7 & 94.5 & 98.0 & 88.8 & 84.6 & 72.8 & 30.9 & 76.4 \\
GraFT (Qwen2.5-VL-7B) & 76.6 & 89.3 & 94.5 & 98.2 & \textbf{88.9} & 88.7 & 45.2 & 30.9 & \textbf{76.9} \\
\midrule
\rowcolor{grouprow}\multicolumn{10}{l}{\textit{Spatial fine-tuned, video input}} \\
ConsiSpace~\cite{consispace} & \textbf{76.6} & \textbf{77.9} & \textbf{78.9} & \textbf{82.2} & 76.3 & 72.1 & \textbf{90.1} & \textbf{54.7} & 79.9 \\
Cambrian-P~\cite{cambrianp} & 73.7 & 74.9 & 60.1 & 76.0 & \textbf{76.9} & \textbf{74.8} & 89.5 & 52.6 & \textbf{85.0} \\
SpaceMind++~\cite{spacemindpp} & 73.4 & 74.5 & 62.4 & 77.9 & \textbf{76.9} & 73.5 & 89.7 & 48.2 & 84.1 \\
\midrule
\rowcolor{grouprow}\multicolumn{10}{l}{\textit{Open-source, video input}} \\
Qwen3.5-4B~\cite{annorollouts} & \textbf{59.7} & 56.5 & 40.4 & 68.9 & 59.3 & 65.5 & 75.7 & 36.1 & \textbf{74.9} \\
Qwen3-VL-8B~\cite{annorollouts} & 57.9 & \textbf{67.5} & \textbf{47.0} & \textbf{76.3} & \textbf{61.9} & 58.0 & 50.9 & 35.0 & 66.3 \\
Qwen3.5-9B~\cite{annorollouts} & 57.9 & 62.0 & 44.0 & 73.6 & \textbf{61.9} & \textbf{69.7} & \textbf{79.9} & \textbf{45.4} & 26.7 \\
\midrule
\rowcolor{grouprow}\multicolumn{10}{l}{\textit{Proprietary, video input}} \\
GPT-6-Astra~\cite{wang2026astra} & \textbf{73.1} & \textbf{61.8} & \textbf{61.5} & \textbf{78.4} & \textbf{56.5} & \textbf{76.1} & \textbf{82.5} & \textbf{87.6} & \textbf{79.9} \\
Gemini-3-Pro~\cite{annorollouts} & 55.1 & 49.0 & 42.8 & 71.5 & 41.8 & 56.6 & 57.5 & 61.9 & 60.0 \\
GPT-5~\cite{annorollouts} & 55.0 & 53.3 & 34.4 & 73.3 & 47.5 & 63.7 & 48.6 & 50.2 & 68.9 \\
Kimi-K2.5~\cite{annorollouts} & 54.5 & 57.2 & 34.9 & 69.3 & 54.4 & 59.6 & 41.3 & 52.1 & 67.0 \\
\bottomrule
\end{tabular}
\caption{GraFT with ground-truth perception on the full VSI-Bench, against the
three strongest published models of each video-input group, plus GPT-6-Astra
on video, whose scores come from a blog evaluation by \citet{wang2026astra}.
GraFT builds the 3DSG from the dataset annotations; the baselines see only the
video, and their scores are as reported in the cited source (harnesses and
frame budgets differ). Among the proprietary rows, GPT-6-Astra was evaluated
at 1 frame per second over the full video (about 117 frames per question),
GPT-5 at 32 frames, and Gemini-3-Pro and Kimi-K2.5 at an unreported budget.
Best per column within each group is in \textbf{bold}.}
\label{tab:upper-bound}
\end{table*}

GraFT reads every answer off the 3DSG, so its accuracy is capped by the
quality of the graph rather than by the backbone. To measure that cap, we
build the graph from the dataset annotations (ground-truth perception,
Appendix~\ref{sec:impl}) and run GraFT unchanged on the full VSI-Bench
(Table~\ref{tab:upper-bound}, Figure~\ref{fig:upper-bound}). With
GPT-6-Astra, GraFT averages $89.5$ (GPT-o3: $88.3$); the four metric subtasks
and both relational subtasks score between $88.8$ and $100.0$. The headroom
left is in route planning ($66.0$) and appearance order ($74.4$), the two
subtasks that also depend on multi-step reasoning or on temporal evidence
beyond the graph, and the only two where a video model scores higher
(GPT-6-Astra, $87.6$; Cambrian-P, $85.0$). For reference, the strongest spatial
fine-tuned model, ConsiSpace, averages $76.6$ ($+12.9$ for GraFT), and
GPT-6-Astra itself reaches $73.1$ from the video alone, $16.4$ points below
GraFT with the same backbone. These gaps are not a like-for-like comparison,
since the baselines see only the video; they show what GraFT reaches once
perception is reliable. Perception is therefore the main bottleneck: on the
same Qwen2.5-VL-7B backbone, the VSI-Bench average rises from $50.0$ under
realistic perception (Table~\ref{tab:vsi-realistic}) to $76.6$ with the
ground-truth graph.

\end{document}